\documentclass[12pt]{article}

\usepackage{graphicx,psfrag,epsf}
\usepackage{enumerate}
\usepackage{natbib}
\usepackage{mathtools}
\usepackage{booktabs}
\usepackage{color}
\usepackage[english]{babel} % English language/hyphenation
\usepackage[protrusion=true,expansion=true]{microtype} % Better typography
\usepackage{amsmath,amsfonts,amsthm}
\usepackage{dsfont}
\usepackage{amssymb}
\usepackage{chngcntr}
\usepackage[toc,page]{appendix}
\usepackage{textcomp}
\usepackage{url}

\usepackage{array}
\usepackage{booktabs} % Horizontal rules in tables
\usepackage{natbib}
\usepackage{setspace}

\usepackage{soul}
\usepackage{multirow}
\usepackage{enumitem}
\usepackage{microtype} % Slightly tweak font spacing for aesthetics
\usepackage[font = small,labelfont=bf,textfont=it]{subcaption}
\usepackage{xcolor}
\usepackage{tabularx}
\usepackage[font = small,labelfont=bf,textfont=it]{caption} % Custom captions under/above floats in tables or figures
\usepackage{footnote}
\usepackage{algorithm}% http://ctan.org/pkg/algorithms
\usepackage[noend]{algpseudocode}% http://ctan.org/pkg/algorithmicx
\usepackage{etoolbox}

\algblock{ParFor}{EndParFor}
\algnewcommand\algorithmicparfor{\textbf{for}}
\algnewcommand\algorithmicpardo{\textbf{do\ parallel}}
\algnewcommand\algorithmicendparfor{\textbf{end\ parallel\ for}}
\algrenewtext{ParFor}[1]{\algorithmicparfor\ #1\ \algorithmicpardo}
\algrenewtext{EndParFor}{\algorithmicendparfor}

\makeatletter
\def\BState{\State\hskip-\ALG@thistlm}

\newcommand{\distas}[1]{\mathbin{\overset{#1}{\kern\z@\sim}}}%
\newcommand{\bm}[1]{\mathbf{#1}}
\newcommand{\bb}[1]{\boldsymbol{#1}}
\newsavebox{\mybox}\newsavebox{\mysim}
\newcommand{\distras}[1]{%
  \savebox{\mybox}{\hbox{\kern3pt$\scriptstyle#1$\kern3pt}}%
  \savebox{\mysim}{\hbox{$\sim$}}%
  \mathbin{\overset{#1}{\kern\z@\resizebox{\wd\mybox}{\ht\mysim}{$\sim$}}}%
}

\newcommand{\be}{\begin{equation}}
\newcommand{\ee}{\end{equation}}
\newcommand{\bi}{\begin{itemize}}
\newcommand{\ei}{\end{itemize}}
\newcommand{\ben}{\begin{enumerate}}
\newcommand{\een}{\end{enumerate}}

\newcommand{\R}{\mathbb{R}}
\newcommand{\D}{\mathcal{D}}
\newcommand{\E}{\mathcal{E}}

\DeclarePairedDelimiter\set\{\}
\DeclarePairedDelimiter{\norm}{\lVert}{\rVert}

\newcolumntype{K}[1]{\geq {\centering\arraybackslash}p{#1}}
\allowdisplaybreaks
\DeclareMathOperator*{\argmin}{\arg\!\min}
\makeatother

\let\oldbibliography\thebibliography
\renewcommand{\thebibliography}[1]{\oldbibliography{#1}
\setlength{\itemsep}{0pt}} %Reducing spacing in the bibliography.

\newcommand{\blind}{0}

\patchcmd{\footnotemark}{\stepcounter{footnote}}{\refstepcounter{footnote}}{}{}
\usepackage{booktabs}
\newcommand{\ra}[1]{\renewcommand{\arraystretch}{#1}}
\usepackage{makecell}
\usepackage[section]{placeins}

\begin{document}

\def\spacingset#1{\renewcommand{\baselinestretch}%
{#1}\small\normalsize} \spacingset{1}

\if1\blind
{
  \title{\bf SPlit: An Optimal Method for Data Splitting}
  \small
  \author{V. Roshan Joseph and Akhil Vakayil}\hspace{.2cm}\\
% }
  \maketitle
} \fi

\if0\blind
{
  \bigskip
  \bigskip
  \bigskip
  \begin{center}
    {\LARGE\bf  SPlit: An Optimal Method for Data Splitting}\vspace{.2cm}\\
    {V. Roshan Joseph and Akhil Vakayil}\vspace{.2cm}\\
    {Stewart School of Industrial and Systems Engineering\\ 
    Georgia Institute of Technology, Atlanta, GA 30332, USA}\vspace{.2cm}\\
\end{center}
  \medskip
} \fi
%\vspace{.5in}
\bigskip

\vspace{-0.5cm}
\begin{abstract}
\noindent In this article we propose an optimal method referred to as \texttt{SPlit} for splitting a dataset into training and testing sets. \texttt{SPlit} is based on the method of Support Points (SP), which was initially developed for finding the optimal representative points of a continuous distribution. We adapt SP for subsampling from a dataset using a sequential nearest neighbor algorithm. We also extend SP to deal with categorical variables so that \texttt{SPlit} can be applied to both regression and classification problems. The implementation of \texttt{SPlit} on real datasets shows substantial improvement in the worst-case testing performance for several modeling methods compared to the commonly used random splitting procedure.
\end{abstract}

\noindent
{\it Keywords: Cross-Validation, Quasi-Monte Carlo, Testing, Training, Validation.}
%\vfill

%\newpage
\spacingset{1.45} % DON'T change the spacing!

\section{Introduction} \label{sec:intro}
For developing statistical and machine learning models, it is common to split the dataset into two parts: training and testing \citep{Stone1974, Hastie2009}. The training part is used for fitting the model, that is, to estimate the unknown parameters in the model. The model is then evaluated for its accuracy using the testing dataset. The reason for doing this is because if we were to use the entire dataset for fitting, the model would overfit the data and can lead to poor predictions in future scenarios. Therefore, holding out a portion of the dataset and testing the model for its performance before deploying it in the field can protect against unexpected issues that can arise due to overfitting.

In this article we consider only datasets where each row is independent of other rows, that is, we will exclude cases such as time series data. The simplest and probably the most common strategy to split such a dataset is to randomly sample a fraction of the dataset. For example, 80\% of the rows of the dataset can be randomly chosen for training and the remaining 20\% can be used for testing. The aim of this article is to propose an optimal strategy to split the dataset. 

\cite{Snee1977} seems to be the first one who has carefully investigated several data splitting strategies. He proposed DUPLEX as the best strategy which was originally developed by Kennard as an improvement to another popular strategy CADEX \citep{Kennard1969}. Over the time, many other methods have been proposed in the literature for data splitting; see, for example the survey in \cite{Reitermanova2010} and the comparative study in \cite{Xu2018}. Some of these methods will be discussed in the next section after proposing a mathematical formulation of the problem.

%To find an optimal way to split the dataset, we first need to formulate the objective of %splitting. Is the objective to create a training set that gives the best estimation of the %model parameters or is it to create a testing set that gives an unbiased evaluation of the %model performance or is to create training and testing sets that share similar %characteristics? Some of these objectives can be conflicting in nature. In this article, %we will propose a mathematical formulation of the problem and develop an optimal procedure %to split the dataset.

%It is generally accepted that the testing set should have similar characteristics as that %of the training set so that the estimated model's performance in the testing set is %similar to that of the training set. However, we are not aware of any optimal procedure %for creating a testing set that is similar in nature to the training set. The aim of this %article is to propose such an optimal procedure.

%so that it can provide an unbiased estimate of the criterion used for fitting the model

It is also common to hold out a portion of the training set for validation. The validation set can be used for fine-tuning the model performance such as for choosing hyper-parameters or regularization parameters in the model. In fact, the training set can be divided into multiple sets and the model can be trained using cross-validation. Our proposed method for optimally splitting the dataset into training and testing can also be used for these purposes by applying the method repeatedly on the training set.

The article is organized as follows. In Section 2, we provide a mathematical formulation of the problem and propose an optimal splitting method called \texttt{SPlit} based on a technique for finding optimal representative points of a distribution known as Support Points \citep{mak2018support}. Support points are defined only for continuous variables. Therefore, we extend the support points methodology to deal with categorical variables in Section 3 so that \texttt{SPlit} can be applied to both regression and classification problems. We apply \texttt{SPlit} on several real datasets in Section 4 and compare its performance with random subsampling.  Some concluding remarks are given in Section 5. 

\section{Methodology} \label{sec:method}
Let $\bm X=(X_1,\ldots,X_p)$ be the $p$ input variables (or features) and $Y$ the output variable. Let $\D=\{(\bm X_i,Y_i)\}_{i=1}^N$ be the dataset in hand. Our aim is to divide $\D$ into two disjoint and mutually exclusive sets: $\D^{train}$ and $\D^{test}$, where the training set $\D^{train}$ contains $N_{train}$ points and testing set $\D^{test}$ contains $N_{test}$ points with $N_{train}+N_{test}=N$.

\subsection{Mathematical Formulation}
Suppose the rows of the dataset are independent realizations from a distribution $F(\bm X,Y)$:
\begin{equation}\label{eq:F}
    (\bm X_i,Y_i)\overset{iid}{\sim} F,\; i=1,\ldots,N.
\end{equation}
Let $g(\bm X;\bb \theta)$ be the prediction model that we would like to fit to the data, where $\bb \theta$ is a set of unknown parameters in the model. The unknown parameters $\bb \theta$ will be estimated by minimizing a loss function $L(Y,g(\bm X;\bb \theta))$. Typical loss functions include squared or absolute error loss. More generally, the negative of the log-likelihood can be used as a loss function. 

In postulating a prediction model $g(\bm X;\bb \theta)$, our hope is that it will be close to the true model $E(Y|\bm X)$ for some value of $\bb \theta$. However, the postulated model could be wrong and there may not exist a true value for $\bb \theta$. Thus it makes sense to try out different possible models on a training set and check their performance on the testing set so that we can identify the model that is closer to the truth. 

The unknown parameters $\bb \theta$ can be estimated from the training set as
\begin{equation}\label{eq:thetahat}
   \hat{\bb \theta}= \underset{\bb \theta}{\textup{Argmin}} \frac{1}{N_{train}}\sum_{i=1}^{N_{train}}L(Y_i^{train},g(\bm X_i^{train};\bb \theta)),
\end{equation}
which is a valid estimator provided that
\begin{equation}\label{eq:condition2}
   (\bm X_i^{train},Y_i^{train})\sim F ,\;i=1,\ldots,N_{train}.
\end{equation}
How should we split the dataset to obtain training and testing sets? We propose that the dataset should be split in such a way that the testing set gives an unbiased and efficient evaluation of the model's performance fitted using the training set.

To quantify the model's performance,  define the generalization error as in \citet[Ch.~7]{Hastie2009} by
\begin{equation}\label{eq:err}
    \E=E_{\bm X,Y}\{ L(Y,g(\bm X;\hat{\bb \theta}))|\D^{train}\},
\end{equation}
where the expectation is taken with respect to a realization $(\bm X,Y)$ from $F$. Note that we do not include the randomness in $\hat{\bb \theta}$ induced by $\D^{train}$ for computing the expectation. 
%Now, because of the independence assumption in (\ref{eq:F}),
%\begin{equation}\label{eq:err}
%    \E=E_{\bm X,Y}\{ L(Y,g(\bm X;\hat{\bb \theta}))\}.
%\end{equation}

We can estimate $\E$ if we have a sample of observations from $F$ that is independent of the training set. We can use the testing set for this purpose. Thus, an estimate of $\E$ can be obtained as
\begin{equation}\label{eq:errtest}
    \widehat{\E}=\frac{1}{N_{test}}\sum_{i=1}^{N_{test}}L(Y_i^{test},g(\bm X_i^{test};\hat{\bb \theta})),
\end{equation}
which works if
\begin{equation}\label{eq:condition}
    (\bm X_i^{test},Y_i^{test})\sim F,\;i=1,\ldots,N_{test}.
\end{equation}
%It is tempting to estimate $\E$ using the training set as
%\begin{equation}\label{eq:errtrain}
%    \overline{\E}=\frac{1}{N_{train}}\sum_{i=1}^{N_{train}}L(Y_i^{train},g(\bm %X_i^{train};\hat{\bb \theta})),
%\end{equation}
%which will also work provided that
%\begin{equation}\label{eq:condition2}
%    (\bm X_i^{train},Y_i^{train})\sim F ,\;i=1,\ldots,N_{train}.
%\end{equation}
%However, because of (\ref{eq:thetahat}), $\overline{\E}$ can severely underestimate the %true generalization error. On the other hand, $\widehat{\E}$ is unaffected by model %estimation/tuning and therefore, it is expected to give an unbiased estimate of the %model's performance. 

%This will happen if the testing set is independent of the training set:
%\begin{equation}\label{eq:indep}
%    \left\{(\bm X_i^{test},Y_i^{test})\right\}_{i=1}^{N_{test}}\indep \left\{(\bm %X_i^{train},Y_i^{train})\right\}_{i=1}^{N_{train}},
%\end{equation}
%which necessitates that these two sets should be disjoint and come from independent and %identically distributed samples as stated in (\ref{eq:F}).
%It is important to note that $\{(\bm X_i^{test},Y_i^{test})\}_{i=1}^{N_{test}}$ need not be an independent sample to make $\hat{\E}$ an unbiased estimator of $\E$. 

A simple way to ensure condition (\ref{eq:condition}) is to randomly sample $N_{test}$ points  from $\D$. Then, (\ref{eq:errtest}) can be viewed as the Monte Carlo (MC) estimate of $\E$. The question we are trying to answer is, if there is a better way to sample from $\D$ so that we can get a more efficient estimate of $\E$. The answer to this question is affirmative. We can use Quasi-Monte Carlo (QMC) methods to improve the estimation of $\E$. It is well known that the error of MC estimates decreases at the rate $\mathcal{O}(1/\sqrt{N_{test}})$, whereas when sampling from uniform distributions, the QMC error rate can be shown to be almost $\mathcal{O}(1/N_{test})$ \citep{niederreiter1992}. This is a substantial improvement in the error rate. However, most QMC methods focus on uniform distributions \citep{Owen2013}. Recently, \cite{mak2018support} developed a method known as \emph{support points} to obtain a QMC sample from general distributions.  Although their theoretical results guarantee a convergence rate faster than MC by only a $\log N_{test}$ factor, much faster convergence rates are observed in practical implementations. This leads us to the proposed method \texttt{SPlit} (stands for Support Points-based split). We will discuss the method of support points and SPlit in detail after reviewing the existing data splitiing methods in the next section.

%A possible drawback of using QMC methods, and in particular support points, to create a testing set is that the independence condition in  (\ref{eq:indep}) may not hold true. Fortunately, the amount of bias it creates seems to be small, which we will show using simulations in a later section.

\subsection{Review of Data Splitting Methods}

\begin{figure}[h]
\begin{center}
\includegraphics[width = 1\textwidth]{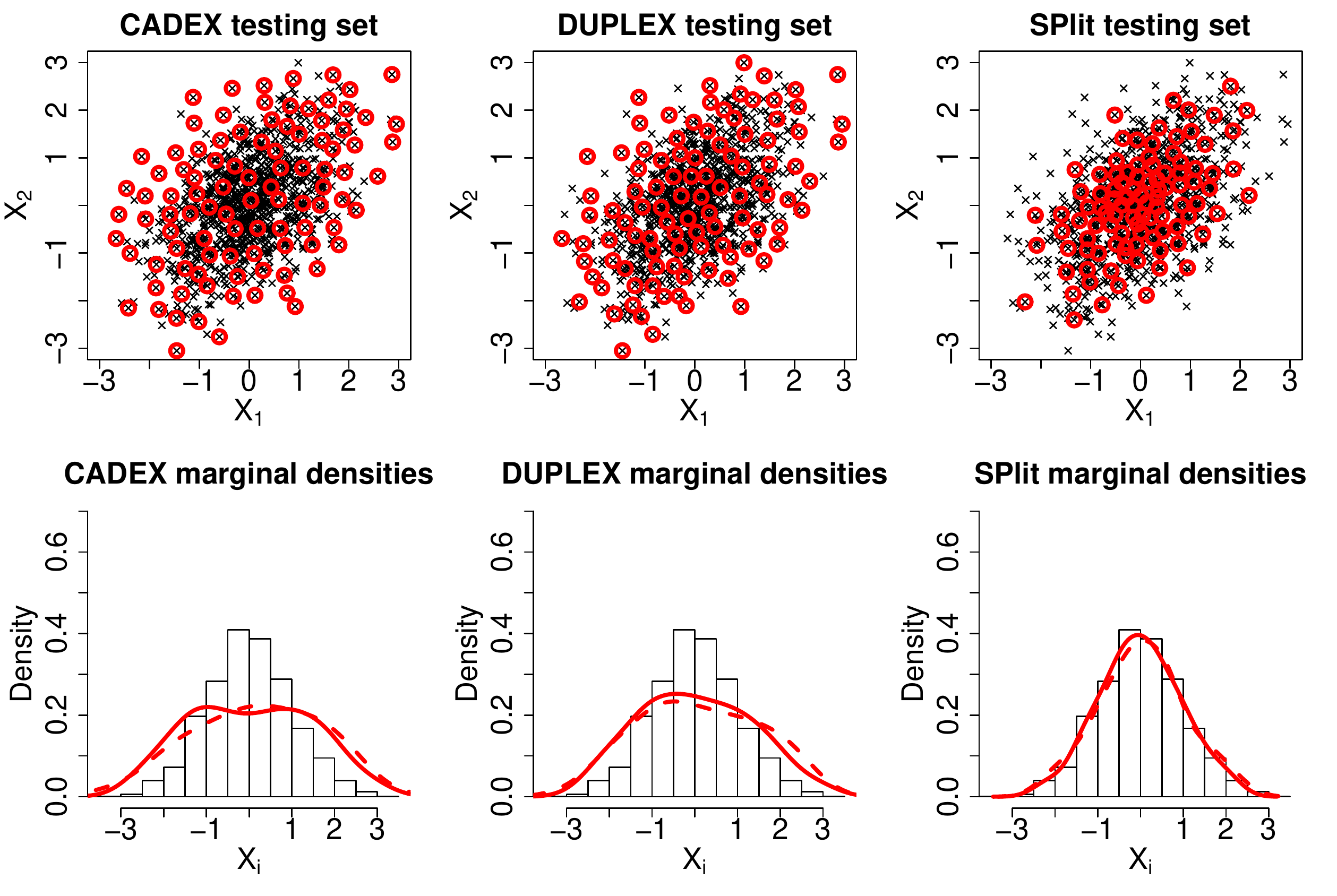} 
\caption{Comparison of CADEX and DUPLEX testing sets with \texttt{SPlit} testing set. 100 testing points (red circles) are chosen from 1,000 data points (black crosses). In the lower panels, the marginal densities of the testing sets are plotted over the histogram of the full data.}
\label{fig:duplex}
\end{center}
\end{figure}

Interestingly, the original motivation behind CADEX \citep{Kennard1969} and DUPLEX \citep{Snee1977} was to create two sets with similar statistical properties, which agrees with the distributional condition mentioned in (\ref{eq:condition}). However, these algorithms cannot achieve this objective. For example, consider a two-dimensional data generated using $(X_{1i},X_{2i})\stackrel{iid}{\sim} N_2(\bm 0,\bm \Sigma)$ for $i=1,\ldots,N$, where $\bm 0=(0,0)'$ and $\Sigma_{jk}=.5^{|j-k|}$. We will omit the response here because these algorithms do not use it. Let $N=1,000$ and $N_{test}=100$. The CADEX and DUPLEX testing sets obtained using the \texttt{R} package \texttt{prospectr} \citep{prospectr} are shown in the left and middle top panels of Figure \ref{fig:duplex}. We can see that both CADEX and DUPLEX testing points are too spread out and therefore, their distributions do not match with the distribution of the data. This can be seen more clearly in the marginal distributions shown in the bottom panels. For comparison, the testing set generated using the proposed \texttt{SPlit} method is shown in the top right panel. We can see that its distribution matches quite well with the distribution of the full data, as desired.

The sample set partitioning based on joint X-Y distances (SPXY) algorithm \citep{Galvao2005} is a modification of the CADEX algorithm which incorporates the distances computed from the response values. Although incorporating $Y$ was in the right direction of (\ref{eq:condition}), the algorithm suffers from the same issues of CADEX and DUPLEX algorithms. \cite{Bowden2002} proposed a data splitting method which uses global optimization techniques to match the mean and standard deviations of the testing set and the full data. This is again in the right direction of (\ref{eq:condition}), however, matching the first two moments does not ensure distributional matching. \cite{May2010} proposed further improvement to the foregoing methodology using clustering-based stratified sampling. Although this provides an improvement, it is well-known that clustering distorts the original distribution \citep{zador1982} and hence cannot satisfy (\ref{eq:condition}). In summary, none of existing data splitting methods except the random subsampling can ensure the distributional condition given in (\ref{eq:condition}) or (\ref{eq:condition2}).

Before proceeding further, we would like to mention about another possible approach to solve the problem. One could think about splitting the dataset in such a way that the training set gives the best possible estimation of the model under a given loss function. For example, it is well-known that the best estimate of a linear regression model with linear effects of the predictors under the least squares criterion can be obtained by choosing the extreme values of the data from the predictor-space \citep{Wangetal2019}. Although such a choice can minimize the variance of the parameter estimates, the model may not perform well in the testing set if the original dataset is not generated from such a linear model. In our opinion, the testing set should provide a set of samples for an unbiased evaluation of the model performance and detect possible model deviations. For example, if we detect that quadratic terms are needed in the linear regression model, then having a training dataset with only extreme values of the predictors is not going to be useful. Therefore, the splitting method should be independent of the modeling choice and the loss function. Random subsampling achieves this aim and we will show that support points will achieve it even better! 

Although support points have been used in the past for subsampling from big data \citep{psp}, it was done for the purpose of saving storage space and time for fitting computationally expensive models due to limited resources. On the other hand, data reduction is not the objective in our problem. After assessing the model's performance using the testing set, the model will be re-estimated using the \emph{full} data before deploying it for future predictions. 

\subsection{Support Points}
Let $\bm Z=(\bm X,Y)$ be a vector of \emph{continuous} variables.  Then, the energy distance between the distribution $F(\bm Z)$ and the empirical distribution of a set of points $\bm z_1,\cdots,\bm z_n$ is defined as \citep{szekely2013energy}
\begin{equation}\label{eq:ED}
   ED=\frac{2}{n} \sum_{i=1}^n \mathbb{E}\|\bm{z}_i - \bm{Z}\|_2 - \frac{1}{n^2} \sum_{i=1}^n \sum_{j=1}^n \|\bm{z}_i - \bm{z}_j\|_2 - \mathbb{E}\|\bm{Z} - \bm{Z'}\|_2,
\end{equation}
where $\bm Z,\bm Z'\sim F$, $\|\cdot\|_2$ is the Euclidean distance, and the expectations are taken with respect to $F$. Note that for the Euclidean distance to make sense,  all the variables are  standardized to have zero mean and unit variance. The energy distance will be small if the empirical distribution of $\bm z_1,\cdots,\bm z_n$ is close to $F$. Therefore, \cite{mak2018support} defined the support points of $F$  as the minimizer of the energy distance:
\begin{equation}
\{\bm z_i^*\}_{i=1}^n \in \underset{\bm{z}_1, \cdots, \bm{z}_n}{\textup{Argmin}} \; {ED} = \underset{\bm{z}_1, \cdots, \bm{z}_n}{\textup{Argmin}} \left\{ \frac{2}{n} \sum_{i=1}^n \mathbb{E}\|\bm{z}_i - \bm{Z}\|_2 - \frac{1}{n^2} \sum_{i=1}^n \sum_{j=1}^n \|\bm{z}_i - \bm{z}_j\|_2 \right\}.
\label{eq:support0}
\end{equation}
They can be viewed as the representative points of the distribution $F$, which is the best set of $n$ points to represent $F$ according to the energy distance criterion. \cite{mak2018support} showed that support points converge in distribution to $F$ and therefore, they can be viewed as a QMC sample from $F$. This property makes support points different from other representative points of a distribution such as MSE-rep points \citep{Fang1994} or principal points \citep{Flury1990} which do not possess distributional convergence \citep{zador1982}.

In our problem, we do not have $F$. Instead we only have a dataset $\D$, which is a set of independent realizations from $F$. Therefore, to compute the support points, we can replace the expectation in (\ref{eq:support0}) with a Monte Carlo average computed over $\D$:
\begin{equation}
\{\bm z_i^*\}_{i=1}^n \in \underset{\bm{z}_1, \cdots, \bm{z}_n}{\textup{Argmin}} \left\{ \frac{2}{nN} \sum_{i=1}^n\sum_{j=1}^N \|\bm{z}_i - \bm{Z}_j\|_2 - \frac{1}{n^2} \sum_{i=1}^n \sum_{j=1}^n \|\bm{z}_i - \bm{z}_j\|_2 \right\}.
\label{eq:support}
\end{equation}
This simple extension to real data settings is another advantage of support points, which cannot be done for other representative points such as minimum energy design \citep{Joseph2015a} and Stein points \citep{Steinpoints2018} even though they possess distributional convergence. 

At first sight, the optimization in (\ref{eq:support}) appears to be a very hard problem. The objective function is nonlinear and non-convex. Moreover, the number of variables in the optimization is $n(p+1)$, which can be extremely high even for small datasets. However, the objective function has a nice feature; it is a difference of two convex functions. By exploiting this feature, \cite{mak2018support} developed an efficient algorithm based on difference-of-convex programming techniques, which can be used for quickly finding the support points. Although a global optimum is not guaranteed, an approximate solution can be obtained in a reasonable amount of time. Their algorithm is implemented in the \texttt{R} package \texttt{support} \citep{supportR}. Although it is possible to create representative points using other goodness-of-fit test statistics \citep{Hickernell1999} and kernel functions \citep{Chen2010} instead of the energy distance criterion, they do not seem to possess the computational advantage and robustness of support points.

\subsection{SPlit} \label{sec:optsub}
We can use the support points obtained from (\ref{eq:support}) as the testing set with $n=N_{test}$ and the remaining data can be used as the training set. Alternatively, we can use support points to obtain the training set with $n=N_{train}$ and then use the remaining data as the testing set. However, the computational complexity of the algorithm used for generating support points is $\mathcal{O}(n^2(p+1))$. Since $N_{test}$ is usually smaller than $N_{train}$, it will be faster to generate the testing set using support points than the training set. In general, we let $n=\min \{N_{test},N_{train}\} = \min \{N\gamma,N(1-\gamma)\}$, where $\gamma=N_{test}/N$ is the splitting ratio.

As discussed earlier, the testing set generated using support points is expected to work better than a random sample from $\D$. However, there is one drawback. Support points need not be a subsample of the original dataset. This is because the optimization in (\ref{eq:support}) is done on a continuous space and therefore, the optimal solution need not be part of $\D$. To get a subsample, we actually need to solve the following \emph{discrete optimization} problem:
\begin{equation}
\{\bm z_i^*\}_{i=1}^{n} \in \underset{\bm{z}_1, \cdots, \bm{z}_{n}\in \D}{\textup{Argmin}} \left\{ \frac{2}{nN} \sum_{i=1}^{n}\sum_{j=1}^N \|\bm{z}_i - \bm{Z}_j\|_2 - \frac{1}{{n^2}} \sum_{i=1}^{n} \sum_{j=1}^{n} \|\bm{z}_i - \bm{z}_j\|_2 \right\}.
\label{eq:supportint}
\end{equation}
Our initial attempts to solve this problem using state-of-the-art integer programming techniques showed that they are accurate, but too slow in finding the optimal solution. Computational speed is crucial for our method to succeed because otherwise it will not be  attractive against the computationally cheap alternative of random subsampling. Therefore, here we propose an approximate but efficient algorithm to subsample from $\D$.

\begin{algorithm}
\caption{\texttt{SPlit}: Splitting a dataset $\D$ with splitting ratio $\gamma$ [\texttt{R} package: \texttt{SPlit}]}
\begin{algorithmic}[1]
\State Input $\D \in \R^{N \times (p+1)}$ and $\gamma =N_{test}/N$ 
\State Standardize the columns of $\D$
\State $n \leftarrow \min\set{N\gamma, N(1-\gamma)}$
%\Comment{$[\cdot]$ is the nearest integer function }
\State Compute $\{\bm z_i^*\}_{i=1}^{n}$ using (\ref{eq:support})
\State $\D_1 \leftarrow \set{}$
\For{$i \in \set{1, \dots, n}$}
    \State $\hat{\bm u} \in  \argmin_{\bm{u}} \set{\norm{\bm u - \bm z_i^*}_2 : \bm u \in \D}$
    %\Comment{If $|\argmin_{\bm{u}} \set{\norm{\bm u - \bm z_i^*}_2 : \bm u \in \D}| > 1$, pick any}
    \State $\D_1 \leftarrow \D_1 \cup \set{\hat{\bm u}}$ 
    \State $\D \leftarrow \D \setminus \set{\hat{\bm u}}$
\EndFor
\State \textbf{end for}
\State $\D_2 \leftarrow \D$
\State \textbf{return} $\D_1, \D_2$
\end{algorithmic}
\end{algorithm}

We will first find the support points in a continuous space as in (\ref{eq:support}), which is very fast. We will then choose the closest points in $\D$ to $\{\bm z_i^*\}_{i=1}^{n}$ according to the Euclidean distance. This can be done efficiently even for big datasets using KD-Tree based nearest neighbor algorithms. However, a naive nearest neighbor assignment can lead to duplicates and therefore, the remaining data points can become more than $N-n$. Moreover, separating the points can increase the second term  in (\ref{eq:supportint}) and thus potentially improve the energy distance criterion. This can be achieved by doing the nearest neighbor assignment sequentially. Our method is summarized in Algorithm 1 and is implemented in the R package \texttt{SPlit}. A critical step in this algorithm is to update the KD-Tree efficiently when a point is removed from the dataset. We use \texttt{nanoflann}, a \texttt{C++} header-only library \citep{blanco2014nanoflann}, which allows for lazy deletion of a data point from the KD-Tree without having to rebuild the KD-Tree every time a point is removed from the dataset.

\subsection{Visualization}

\begin{figure}[h]
\begin{center}
\begin{tabular}{cc}
\includegraphics[width = 0.45\textwidth]{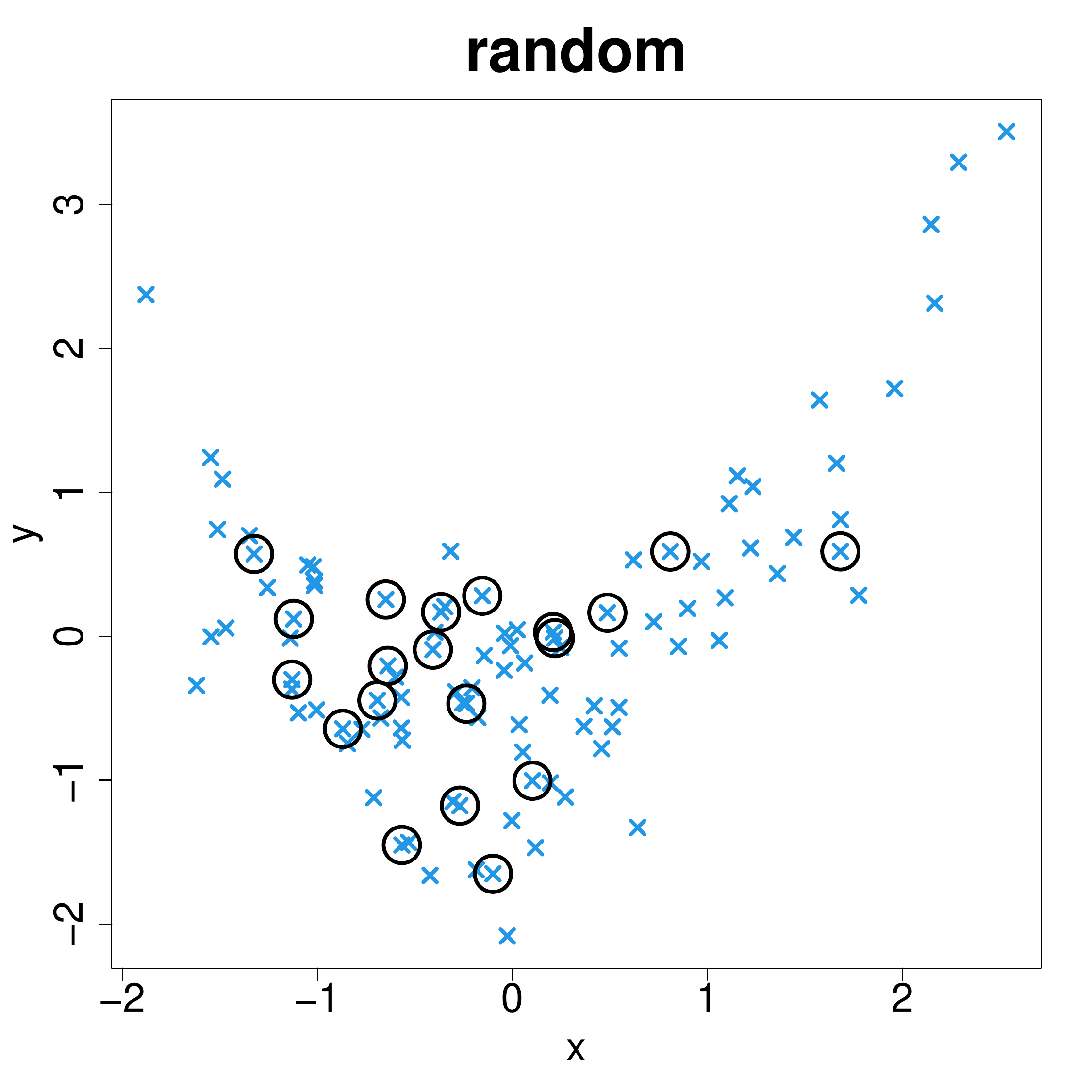} &
\includegraphics[width = 0.45\textwidth]{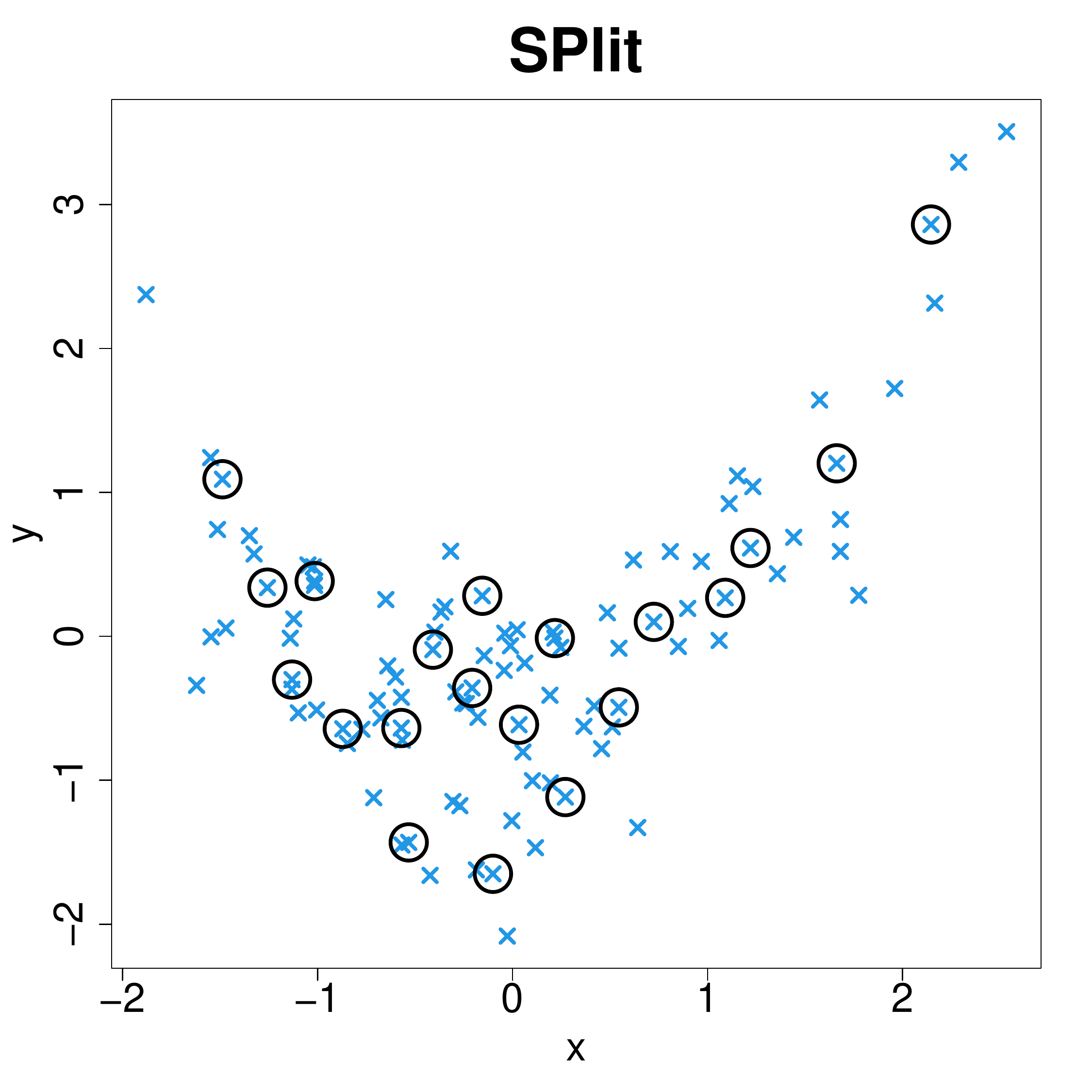}
\end{tabular}
\caption{The circles are the testing set obtained using random (left) and \texttt{SPlit} (right) subsampling.}
\label{fig:visual}
\end{center}
\end{figure}

Consider a simple example for visualization purposes. Suppose we generate $N=100$ points as follows: $X_i\overset{iid}{\sim} N(0,1)$ and $Y_i|X_i\overset{iid}{\sim} N(X_i^2,1)$ for $i=1,\ldots,N$. Both $X$ and $Y$ values are standardized to have zero mean and unit variance. Figure \ref{fig:visual} shows the optimal testing set obtained using \texttt{SPlit} and a random testing set obtained using random subsampling without replacement. We can see that the points in the \texttt{SPlit} testing set are well spread out throughout the region and provide a much better point set to evaluate the model performance than the random testing set.

Most statistical and machine learning models have some hyper-parameters or regularization parameters, which are commonly estimated from the training set by holding out a validation set, say of size $N_{valid}$. One simple approach to create an optimal validation set is to apply the \texttt{SPlit} algorithm on the training set. However, it may happen that such a set is close to the points in the testing set, which is not good as it may lead to a biased testing performance. We want the validation points to stay away from the testing points so that the testing performance is not influenced by the model estimation/validation step. This can be achieved as follows. Let $\{\bm z_1,\ldots,\bm z_{N_{test}}\}$ be the testing set and $\{\bm z_{N_{test}+1},\ldots,\bm z_n\}$ the validation set, where $n=N_{test}+N_{valid}$. Then, the optimal validation points can be obtained as
\begin{equation}
\{\bm z_i^*\}_{i=N_{test}+1}^{n} \in \underset{\bm{z}_{N_{test}+1}, \cdots, \bm{z}_{n}\in \D}{\textup{Argmin}} \left\{ \frac{2}{nN} \sum_{i=1}^{n}\sum_{j=1}^N \|\bm{z}_i - \bm{Z}_j\|_2 - \frac{1}{{n^2}} \sum_{i=1}^{n} \sum_{j=1}^{n} \|\bm{z}_i - \bm{z}_j\|_2 \right\},
\label{eq:valid}
\end{equation}
where the optimization takes place only over the validation points with the testing points fixed at $\{\bm z_1^*,\ldots,\bm z_{N_{test}}^*\}$. Because of the second term in the energy distance criterion, the validation points will move away from the testing points.

\begin{figure}[h]
\begin{center}
\begin{tabular}{cc}
\includegraphics[width = 0.45\textwidth]{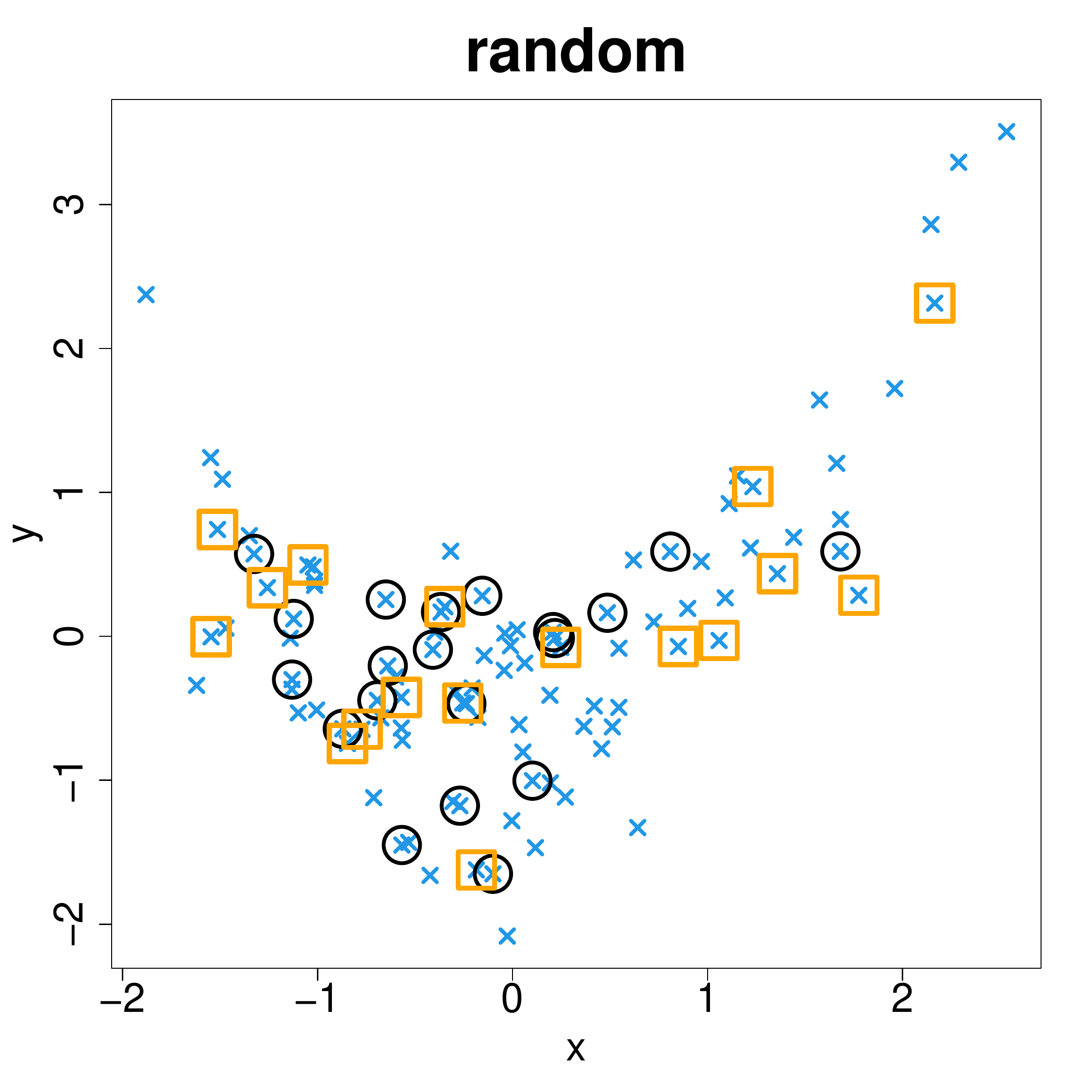} &
\includegraphics[width = 0.45\textwidth]{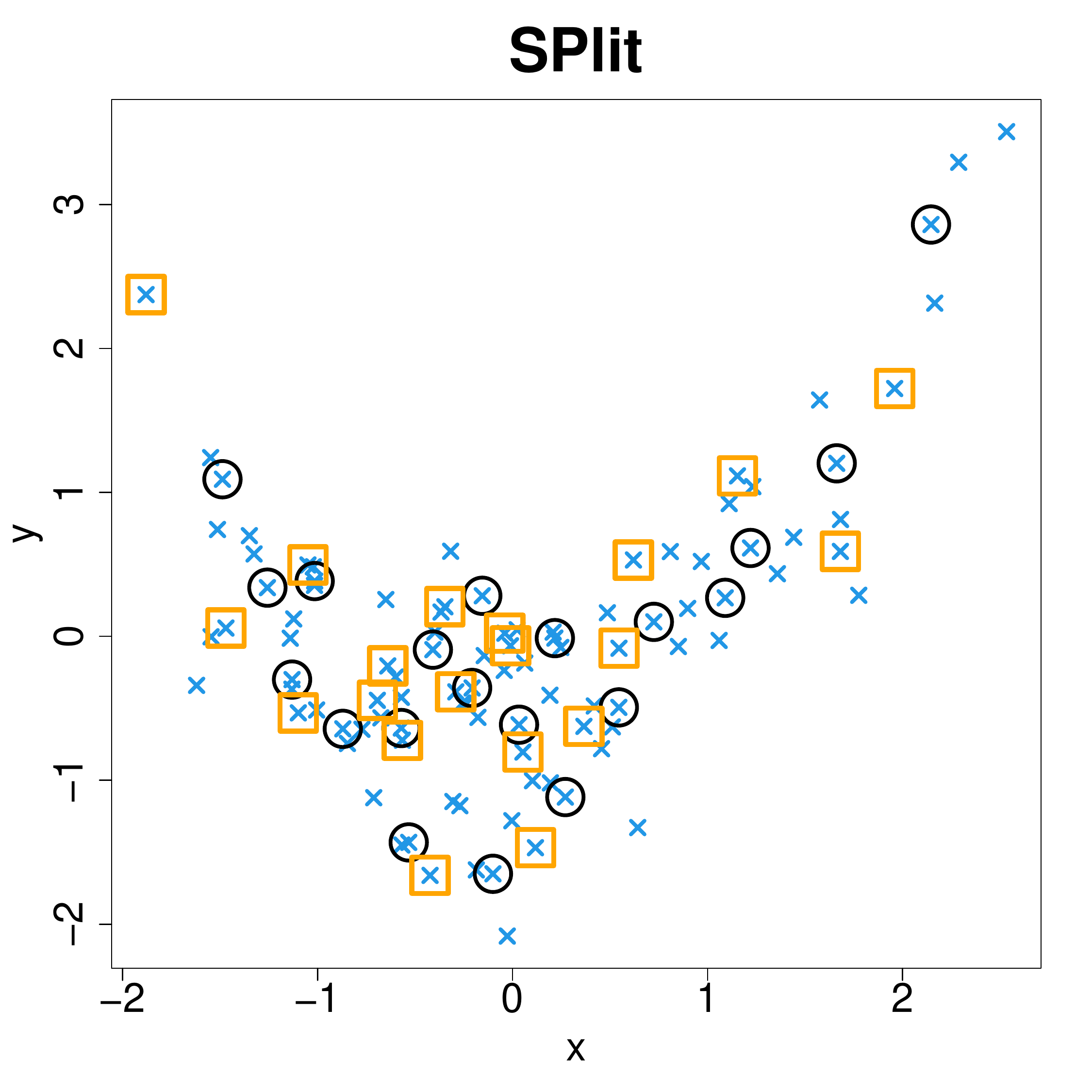}
\end{tabular}
\caption{The squares are the validation set obtained using random (left) and \texttt{SPlit} (right) subsampling from the training set. The testing set is shown as circles.}
\label{fig:visual2}
\end{center}
\end{figure}

Figure \ref{fig:visual2} shows 20 points selected out of the 80 training points using (\ref{eq:valid}). A random subsample is also shown in the same figure for comparison. Clearly, the optimal validation set created using our method is a much better representative set of the original dataset and therefore, it can do a much better job in tuning the hyper-parameter or regularization parameter than using a random validation set. In fact, we can sequentially divide the training set into $K$ sets by repeated application of this method and use them for $K$-fold cross-validation. Because of the importance of this problem, we will leave this topic for future research. 

\subsection{Simulations}
Since support points create a dependent set, one may wonder if the testing set and training set are related and if it will create a bias in the estimation of the generalization error in (\ref{eq:errtest}). We will perform some simulations to check this. Consider again the data generating model discussed in the previous section:
\begin{equation}\label{eq:sim_model}
    Y_i= X_i^2+\epsilon_i,
\end{equation}
where $\epsilon_i\overset{iid}{\sim} N(0,1)$ and $X_i\overset{iid}{\sim} N(0,1)$, for $i=1,\ldots,N$. Let $N=1,000$. Suppose we fit the following $r$th degree polynomial model to the data:
\[Y_i=g(X_i;\bb \theta)+\epsilon_i,\]
where $g(X;\bb \theta)=\theta_0+\theta_1 X+\theta_2 X^2+\cdots +\theta_r X^r$ and $\epsilon_i\overset{iid}{\sim} N(0,\sigma^2)$. The unknown parameters $\bb \theta=(\theta_0,\theta_1,\ldots,\theta_r)'$ can be estimated from the training set using least squares. The generalization error can then be computed as
\begin{eqnarray*}
\E&=&E_{X,Y}\left[ \{Y-g(X;\hat{\theta})\}^2|\D^{train}\right]\\
&=&E_{X,Y}\left[ \{Y-g(X;\hat{\theta})\}^2\right] \quad \text{(by independence)}\\
&=&E_{X}\left(E_{Y|X}\left[ \{Y-g(X;\hat{\theta})\}^2|X\right]\right)\\
&=&E_{X}\left(\{X^2-g(X;\hat{\bb \theta})\}^2\right)+1.
\end{eqnarray*}

\begin{figure}[h]
\centering
\includegraphics[width=\columnwidth]{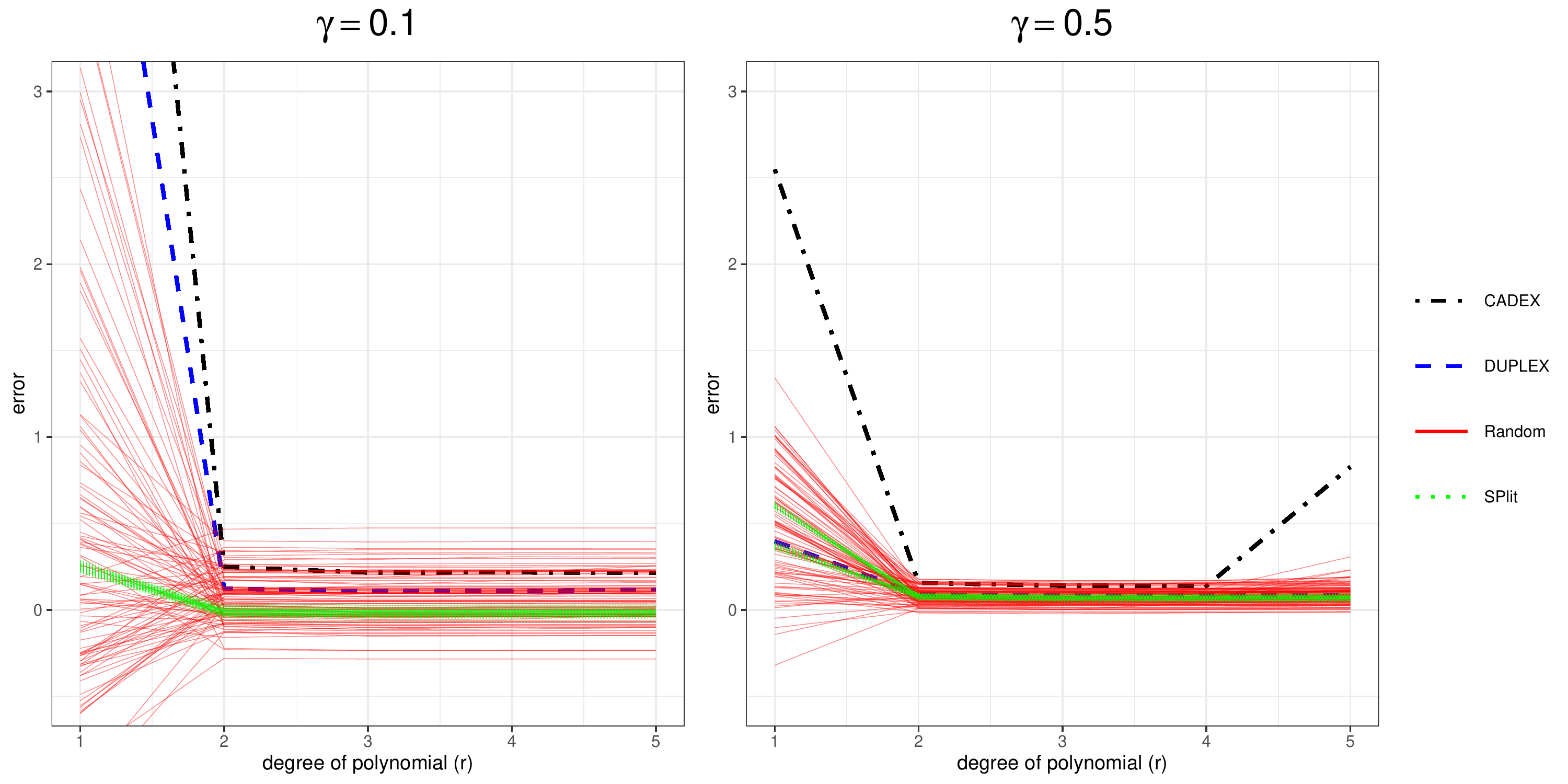}
\caption{The error in the estimation of generalization error with splitting ratios of 10\% (left) and 50\% (right) for different data splitting methods over 100 iterations.}
\label{fig:error}
\end{figure}

We can divide a given dataset into training and testing sets using various data splitting methods and estimate the generalization error using (\ref{eq:errtest}). Thus, we can compute the estimation error of a data splitting method as $\hat{\E}-\E$. For comparison, we use SPlit, random subsampling, CADEX, and DUPLEX. This procedure is repeated 100 times by generating testing sets with splitting ratios of 10\% and 50\%. Owing to the deterministic nature, CADEX and DUPLEX produce the same testing set each time. On the other hand, some variability is observed in the testing sets produced by SPlit, which is mainly due to the random initialization and convergence to local optima of the support points' algorithm. Figure \ref{fig:error} shows the estimation errors over different values of $r$.

% \begin{figure}[h]
% \begin{center}
% \begin{tabular}{cc}
% \includegraphics[width = 0.45\textwidth]{error10.pdf} &
% \includegraphics[width = 0.45\textwidth]{error50.pdf}
% \end{tabular}
% \caption{The error in the estimation of generalization error with splitting ratios of 10\% (left) and 50\% (right) for different data splitting methods.}
% \label{fig:error}
% \end{center}
% \end{figure}

We can see that the bias in the estimation of generalization error using SPlit is small compared to the other data splitting methods. This confirms the validity of the proposed method.

\section{Categorical Variables}
Energy distance in (\ref{eq:ED}) is defined only for continuous variables because its definition involves Euclidean distances and therefore, support points can only be found for datasets with continuous variables. However, a dataset can have categorical predictors and/or responses. Therefore, it is important to extend the support points methodology to deal with categorical variables in order to implement the \texttt{SPlit}.

For simplicity of notations, let us consider the case of only a single categorical variable, which could be a predictor or a response. It is easy to extend our methodology to multiple categorical variables, which will be explained later. Let $m$ be the number of levels of the categorical variable and $N_i$ be the corresponding number of points in the dataset at the $i$th level, $i=1,\ldots,m$. The most naive approach to deal with a categorical variable is to simply ignore it and find the $n$ support points from a dataset containing $N$ points using only the continuous variables as in (\ref{eq:support}).  Suppose we use the nearest neighbor algorithm described in Section 2.3 to select a subsample from the dataset. We can count the number of subsamples with the $i$th level of the categorical variable. Denote it by $n_i$. This subsample gives a good representation of the categorical variable in the dataset if \begin{equation}\label{eq:propsamp}
\frac{n_i}{n}\approx \frac{N_i}{N} \;\;\textrm{for all}\;\; i=1,\ldots,m.
\end{equation}
This is likely to happen even with the naive approach, but it is not guaranteed. For example, the points in the continuous space need not be unique or some of them may be very close to each other and therefore, the nearest neighbor assignment need not balance the levels of the categorical variable.

Another possible approach that can ensure the proportional sampling in (\ref{eq:propsamp}) is to first choose $n_i\approx N_i/Nn$ and then find $n_i$ support points from the dataset containing only the $i$th level of the categorical variable while ignoring the remaining part of the dataset. Although this stratified proportional sampling approach ensures perfect balancing of categorical levels, the support points in the continuous space may not be representative. For example, the support points from different levels of the categorical variable can project onto the same points and thus not space-filling in the continuous variable-space.

A yet another approach to deal with categorical variables is to first convert them into numerical variables and then use the methodology that we developed for continuous variables. Such an approach can be easily adapted for both nominal and ordinal variables. Moreover, it can be easily used for multiple categorical variables, by converting each categorical variable into numerical variables. Thus, this approach appears to be very general and simple to implement. Therefore, we will adopt this idea here, the details of which are developed in the next subsection.

\subsection{Coding}
\label{sec:coding}
If the categorical variable is of ordinal type, then we can numerically score them and convert it into a continuous variable \citep[p.~647]{wu-hamada2011}. So we only need to deal with the case of nominal type variables. If the nominal categorical variable has $m$ levels, then the first step is to represent it using $m-1$ dummy variables assuming that the model has a parameter to represent the mean of the data.

The most popular method to create dummy variables is the treatment coding (also known by the names dummy-level coding and one-hot coding). First create $m$ dummy variables, where the $i$th dummy variable takes the value 1 if the categorical variable has level $i$ and 0 otherwise. Now we can remove any one of the $m$ dummy variables giving us the desired $m-1$ dummy variables. There are many other choices for coding such as Helmert coding, sum coding, orthogonal polynomial coding, etc. \citep[ch.14]{faraway2015}. In terms of modeling and prediction, all these codings are equivalent. However, surprisingly, the coding has a big effect in the support points methodology.

\begin{figure}[h]
\centering
\includegraphics[width=\columnwidth]{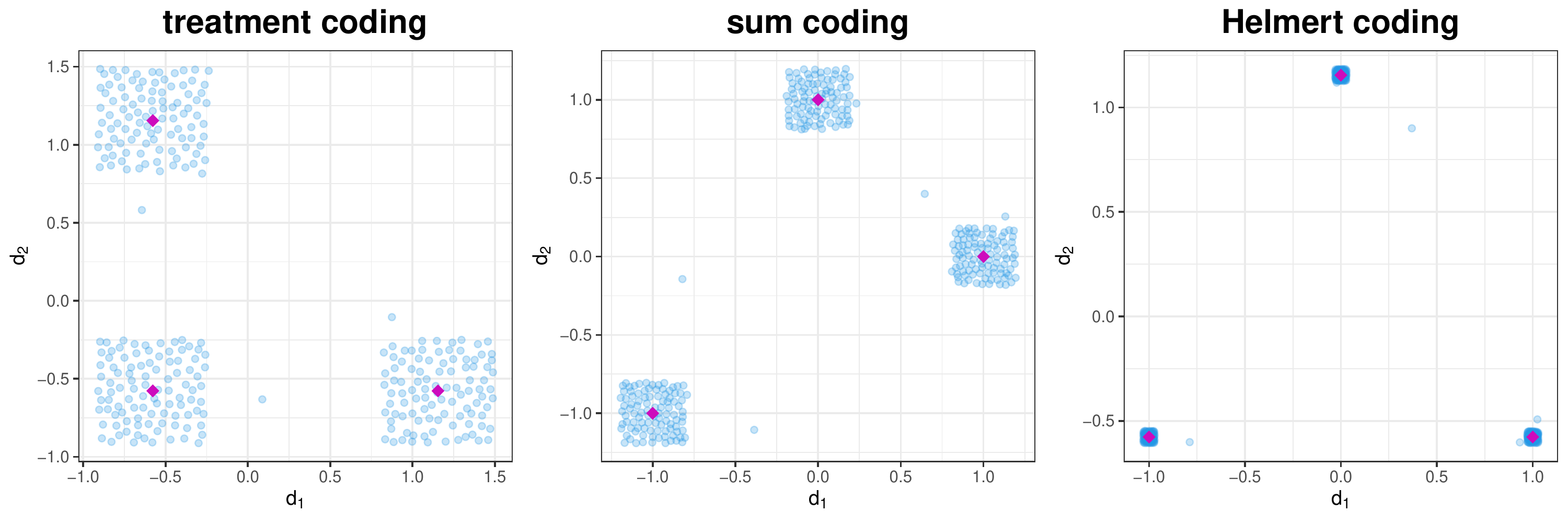}
\caption{Plot of 300 support points obtained from 3,000 points with one categorical variable having three levels using three coding schemes.}
\label{fig:cat3}
\end{figure}

Consider, for example, a categorical variable with three levels. This variable can be represented using two dummy variables: $d_1$ and $d_2$. For treatment coding, $d_1=(0,1,0)$ and $d_2=(0,0,1)$. We will standardize all the variables to have zero mean and unit variance. Then, the dummy variable values become $(-1,2, -1)/\sqrt{3}$ and $(-1,-1, 2)/\sqrt{3}$. Suppose the three levels are replicated 1,000 times creating a dataset of size 3,000. Now we find $n=300$ support points by treating the two dummy variables as continuous variables. The support points are shown in the left panel of Figure \ref{fig:cat3}. We can see that the support points deviate quite a lot from the three target points $\{(-1,-1)/\sqrt{3}, (2,-1)/\sqrt{3}, (-1,2)/\sqrt{3}\}$ (shown as red diamonds in the same figure). This is because of several reasons. First, we are solving the continuous optimization problem in (\ref{eq:support}) and not the discrete optimization problem in (\ref{eq:supportint}). Second, the difference-of-convex program used in solving the optimization problem (\ref{eq:support}) can converge to a local optimum depending on the initialization. Third, the second term in the energy design criterion tries to push the points away from each other. 

The foregoing exercise is repeated using sum coding and Helmert coding. For sum coding, the two dummy variables are coded as $d_1=(1,0,-1)$ and $d_2=(0,1,-1)$ and for Helmert coding they are  $d_1=(-1,1,0)$ and $d_2=(-1,-1,2)$. Standardization doesn't alter the sum coding, whereas $d_2$ in the Helmert coding needs to be divided by $\sqrt{3}$. The 300 support points are shown in the middle and last panels of Figure \ref{fig:cat3} for the sum and Helmert coding, respectively. We can see that the deviation of the support points from the three desired locations are smaller with sum coding compared to treatment coding and even smaller for Helmert coding! This clearly shows that the coding has a big effect on the support points.

% \begin{figure}[h]
% \begin{center}
% \begin{tabular}{ccc}
% \includegraphics[width = 0.3\textwidth]{treatment.pdf} &
% \includegraphics[width = 0.3\textwidth]{sum.pdf}&
% \includegraphics[width = 0.3\textwidth]{helmert.pdf}
% \end{tabular}
% \caption{Plot of 300 support points obtained from 3,000 points with one categorical variable having three levels using three coding schemes.}
% \label{fig:cat3}
% \end{center}
% \end{figure}

Once we obtain the support points, we will use a nearest neighbor algorithm to assign each of the support points to one of the three desired locations (shown as red diamonds in Figure \ref{fig:cat3}). We can expect the nearest neighbor algorithm to perform consistently well and give a balanced allocation if the deviation of support points from these three points are small. In this particular case, the support points from all the three codings form three disjoint clusters and therefore, all of them should perform well. However, in higher dimensions, there can be more deviations and overlaps. Furthermore, in real problems, we will have both continuous and categorical variables; so the deviations are also affected by the values taken by the continuous variables. Therefore, we can expect a more robust performance if the cluster of points are tightly centered around the desired locations. Thus, Helmert coding seems to be the best choice among the three options.

\begin{figure}[h]
\begin{center}
\includegraphics[width = 0.5\textwidth]{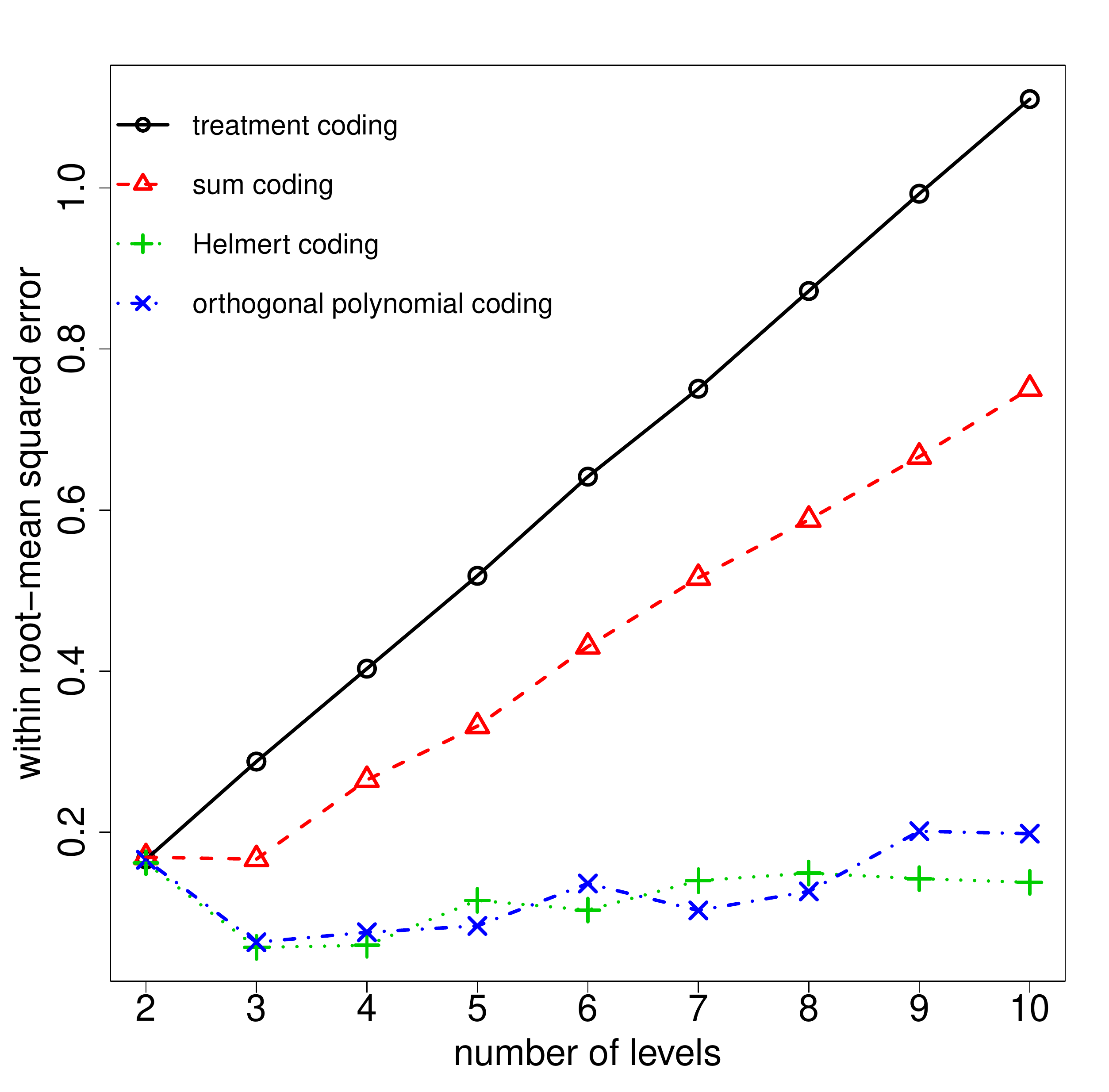}
\caption{Plot of within root-mean squared errors against number of levels of the categorical variable for different coding schemes.}
\label{fig:mse}
\end{center}
\end{figure}

Now consider the case with more than three levels for the categorical variable. As before, we replicate each of the $m$ levels of the categorical variable 1,000 times and then obtain $100m$ support points using four types of coding: treatment, sum, Helmert, and orthogonal polynomial. We have now included orthogonal polynomial coding also in our comparisons because it is different from Helmert coding when $m\ge 4$. We then computed the within cluster sum of squares of deviations (or errors) using the $m$ points specified by the coding as the cluster centers. Figure \ref{fig:mse} shows the within root-mean squared errors for $m=2,3,\ldots,10$. We can see that treatment and sum codings are uniformly worse than both the Helmert and orthogonal polynomial codings.  Between Helmert and orthogonal polynomial codings, there does not seem to be a clear winner.

\begin{figure}[h]
\begin{center}
\begin{tabular}{cc}
\includegraphics[width = 0.45\textwidth]{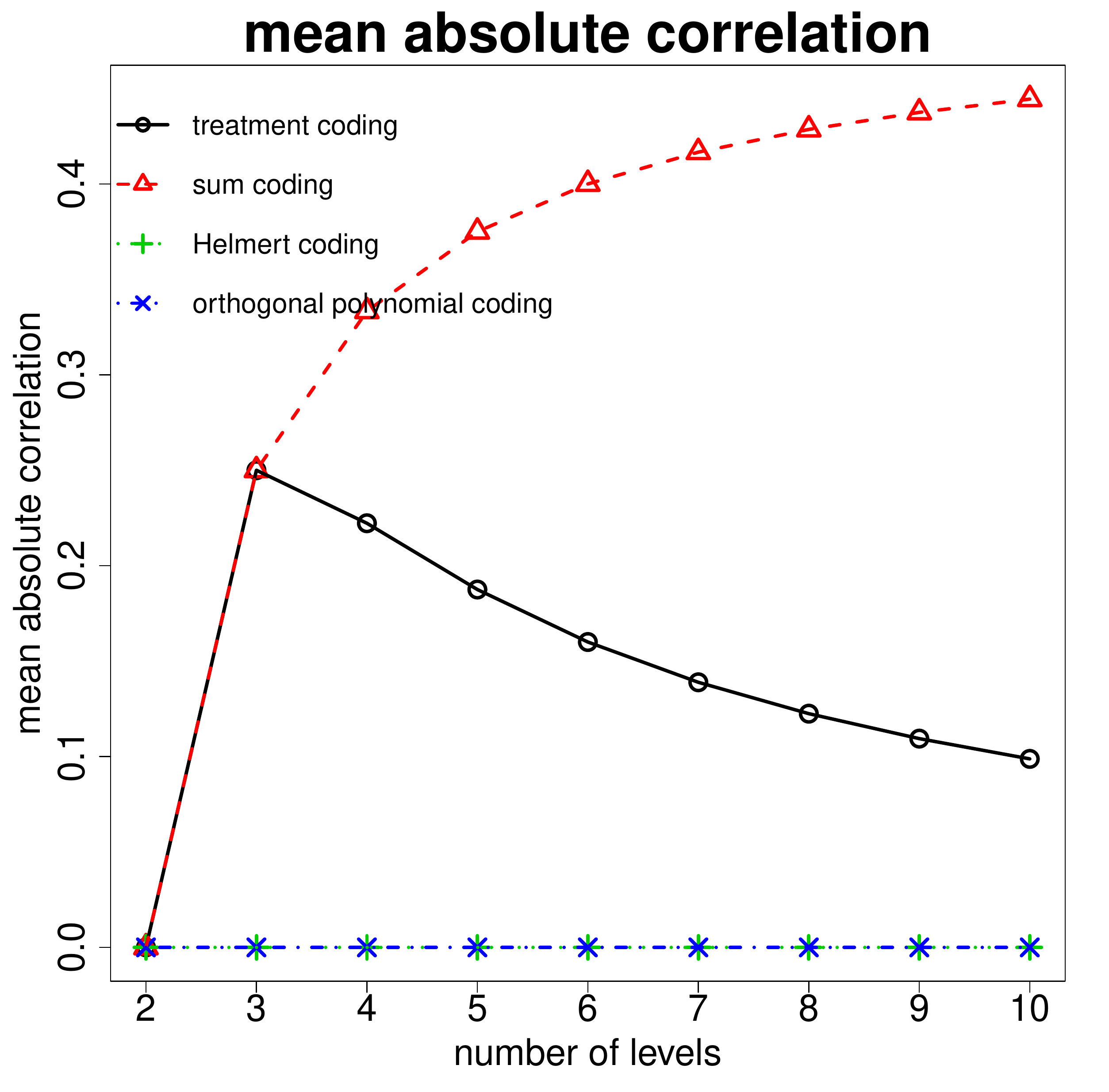} &
\includegraphics[width = 0.45\textwidth]{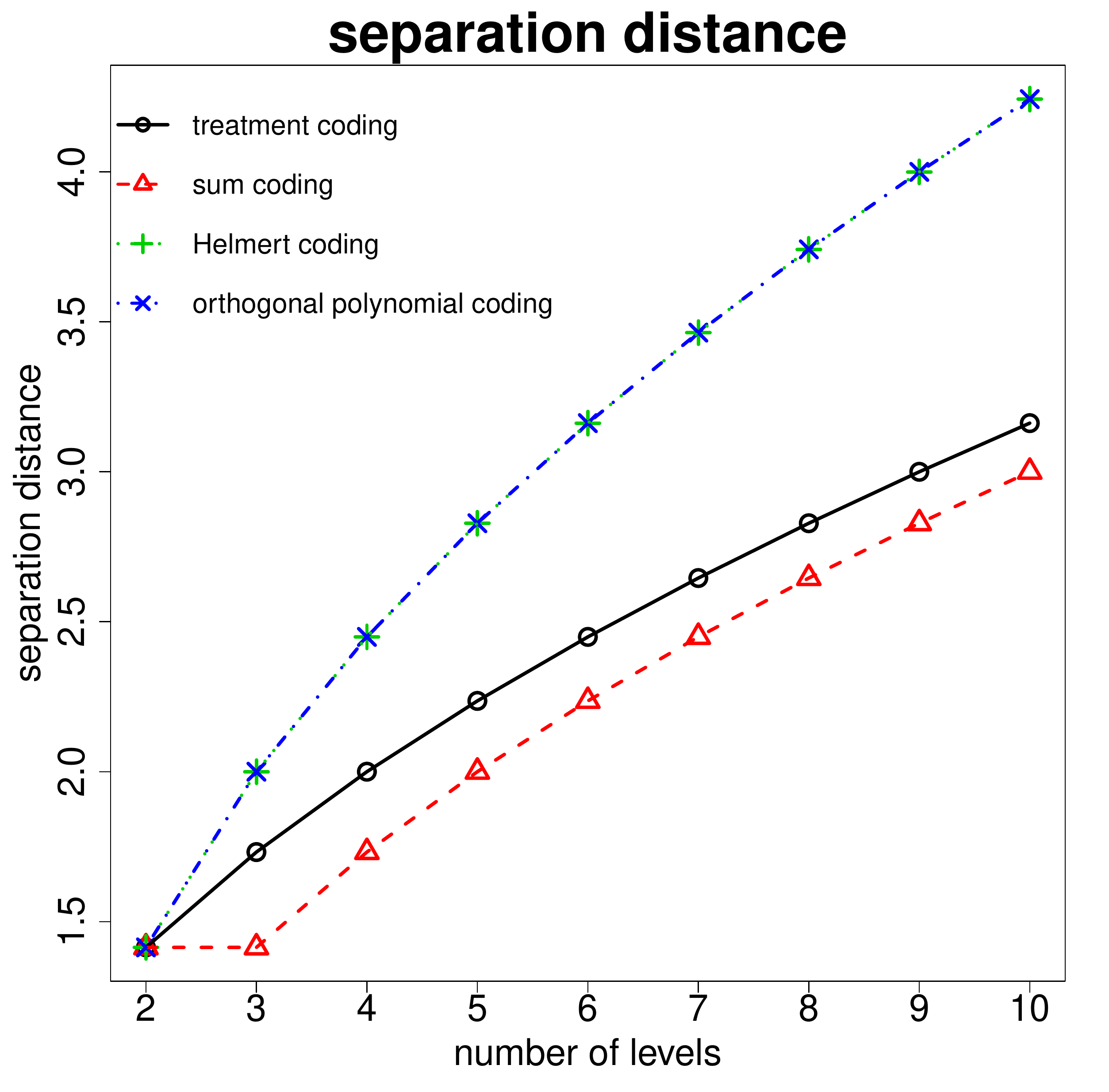}
\end{tabular}
\caption{Mean absolute correlation (left) and separation distance (right) against the number of levels for different coding schemes.}
\label{fig:cor-dist}
\end{center}
\end{figure}

What could be the reason behind this interesting phenomenon? Let $r_{ij}$ be the correlation between the $i^{th}$ and $j^{th}$ column of the coding matrix, which has $m$ rows and $m-1$ columns. The left panel of Figure \ref{fig:cor-dist} shows the average absolute correlation: $\sum_{i\ne j} |r_{ij}|/\{(m-1)(m-2)\}$. We can see that Helmert and orthogonal polynomial codings have zero correlations, whereas the average absolute correlations are high for both the sum and treatment codings. Although this seems to have some connections to the phenomenon observed in Figure \ref{fig:mse}, it does not explain the increase in the treatment coding root-mean squared errors with the increase in the number of levels. The right panel of Figure \ref{fig:cor-dist} shows the separation distance between the $m$ points (minimum distance among all the pairwise distances). We want the separation distance to be as high as possible so that the errors in a nearest neighbor assignment can be minimized. This is similar to a maximin distance criterion used in space-filling designs \citep{Johnson1990}. We can see that the separation distances of both Helmert and polynomial codings are higher than those of treatment and sum codings, which explain the superior performance of the former. However, separation distances do not seem to explain why sum coding is better than treatment coding.

Our aim is to find a coding that minimizes the maximum error from the cluster center of the support points, which seems more closely related to the objective of minimax distance designs \citep{Johnson1990}. However, we do not have a pre-specified hypercube to compute the minimax distance criterion. Thus, we are not able to evaluate the coding schemes using the minimax criterion. However, this does seem to indicate that a dense packing of $m$ spheres in an $m-1$ dimensional space can give us the best coding. This can be achieved using simplex packing of the spheres. Interestingly, both Helmert and orthogonal polynomial codings are rotations of a simplex design. This seems to justify their good performance. It turns out that we can find an optimal rotation of the simplex design as in \cite{He2017} to further improve the root-mean squared error for a given number of levels of the categorical variable. However, the cloud of points in Figure \ref{fig:mse} looks like boxes and not spheres and therefore, we leave this as a topic for future research for a more in-depth investigation. For the rest of the article, we will use Helmert coding for a nominal categorical variable.

\subsection{Visualization}
For illustration, consider the same example used in Section 2.4, except that we generate the data as follows. Let $X_{1i}\overset{iid}{\sim} N(0,1)$ and $X_{2i}|X_{1i}\overset{iid}{\sim} N(X_{1i}^2,1)$ for $i=1,\ldots,N$ with $N=100$. They are scaled to have zero mean and unit variance. Consider a nominal categorical response variable with three levels: Red, Green, and Blue. They are generated using the rule: Red if $6X_1+X_2+6<0$, Blue if $-6X_1+X_2+6<0$, and Green otherwise.

\begin{figure}[h]
\begin{center}
\begin{tabular}{ccc}
\includegraphics[width = 0.31\textwidth]{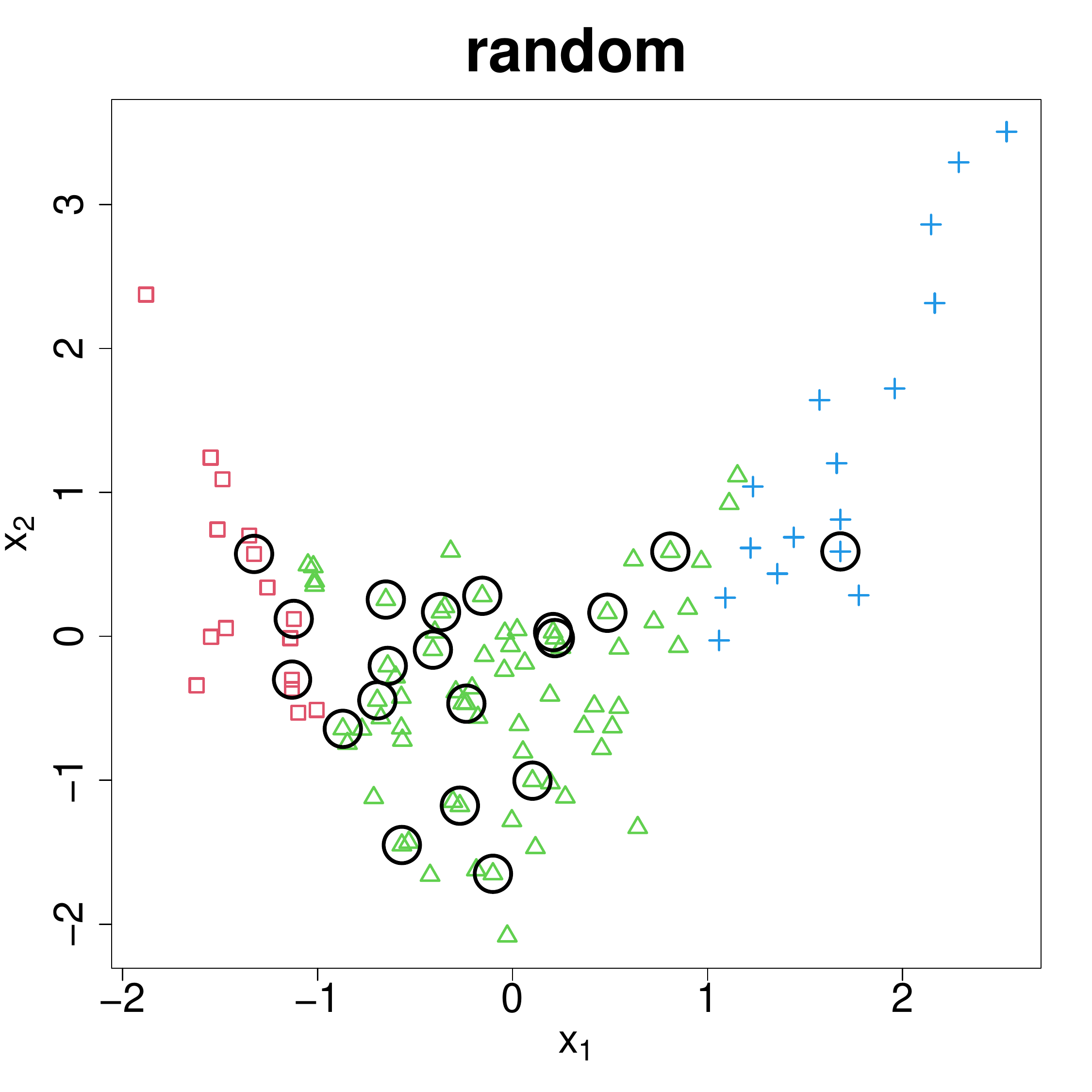} &
\includegraphics[width = 0.31\textwidth]{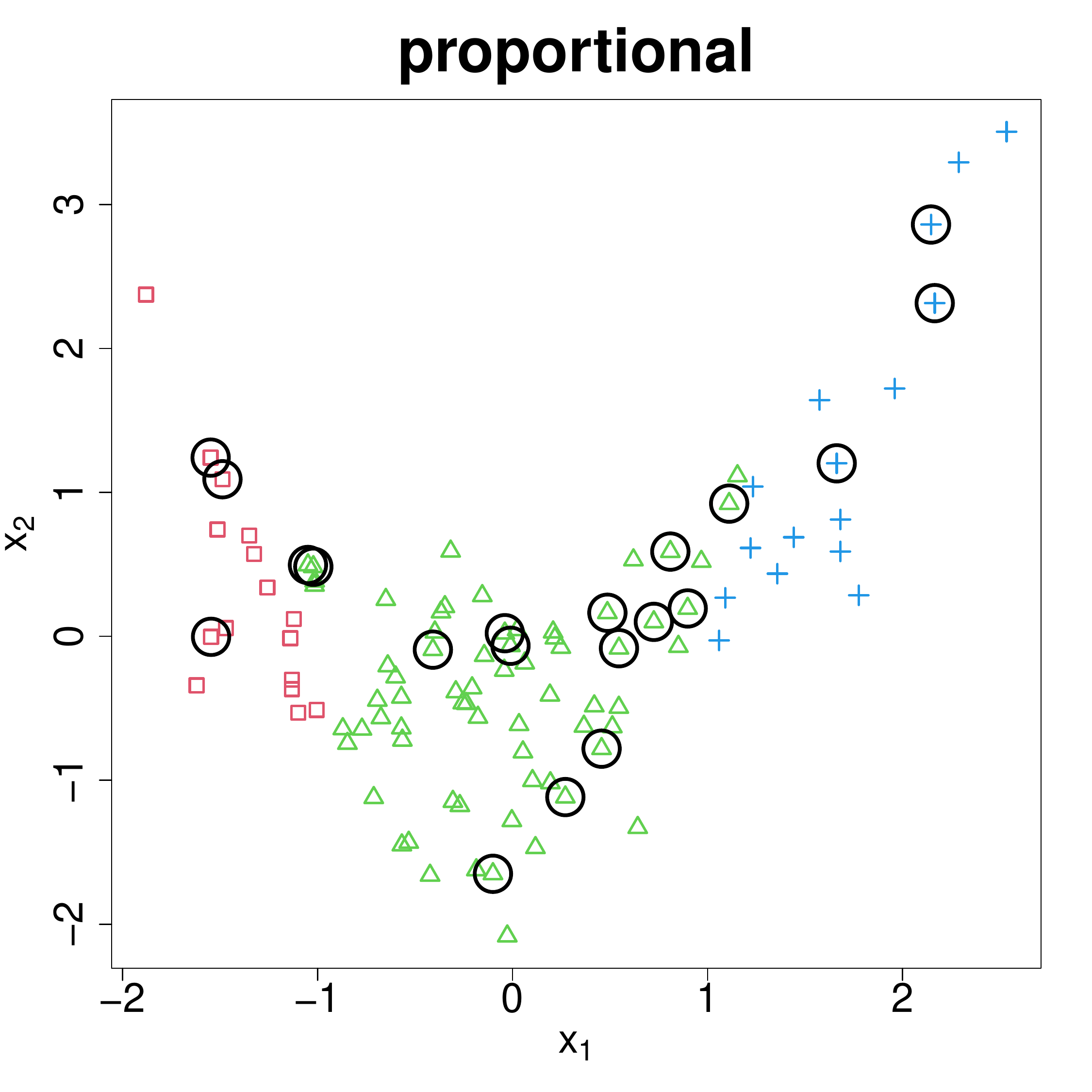} &
\includegraphics[width = 0.31\textwidth]{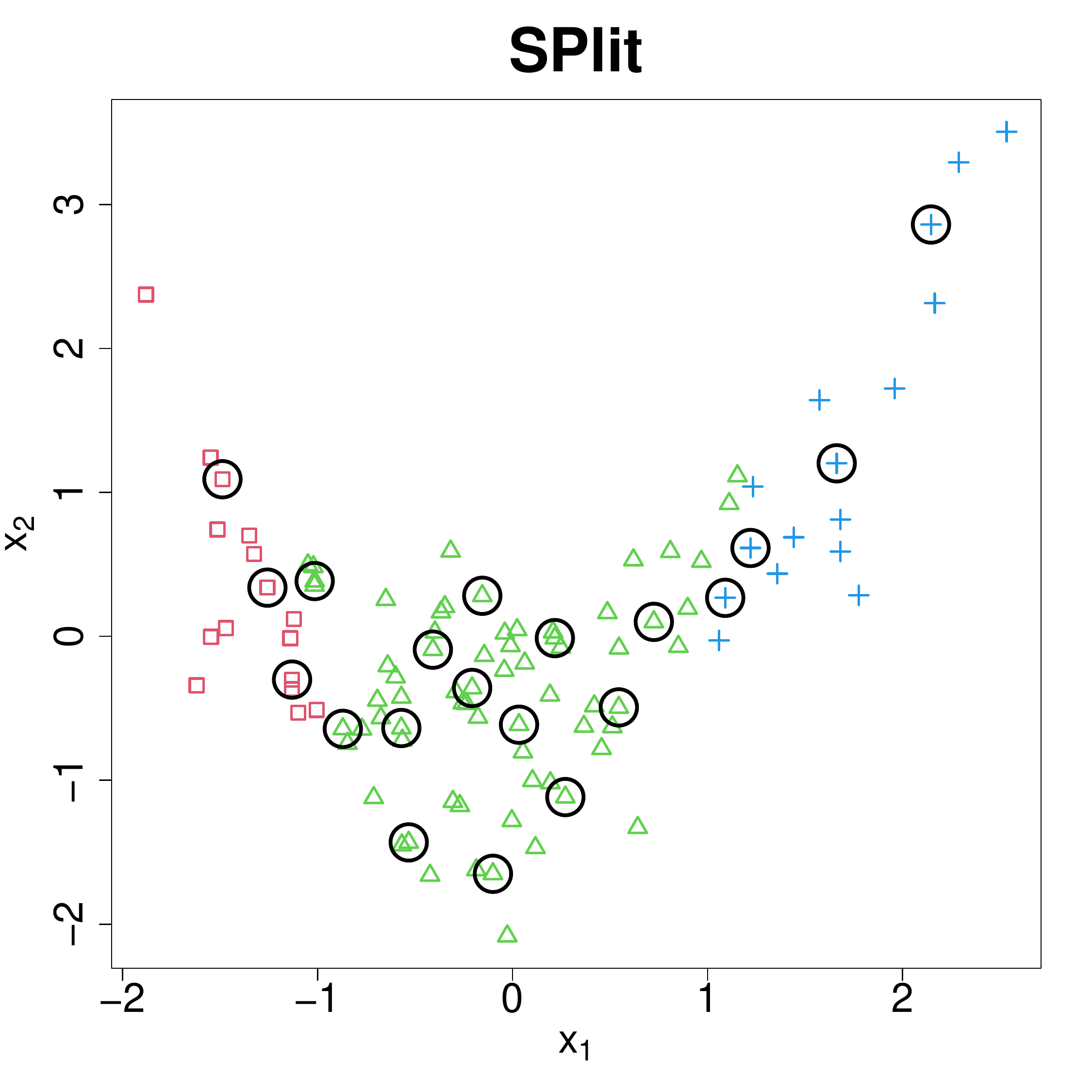}
\end{tabular}
\caption{The circles are the testing set obtained using random (left), stratified proportional (center), and  \texttt{SPlit} (right) subsampling from a dataset with a categorical response variable which has three levels, shown as red squares, green triangles, and blue pluses.}
\label{fig:visual3}
\end{center}
\end{figure}

The left panel of Figure \ref{fig:visual3} shows a random testing set of size $N_{test}=20$. There are 16 Red, 68 Green, and 16 Blue in the original dataset. However, this particular random subsample contains only two Red and one Blue, which is quite disproportionate with respect to the data distribution. We can easily overcome this issue using stratified proportional sampling, that is, by randomly sampling $n_i\approx nN_i/N$ samples from each level $i=1,2,3$. This is shown in the middle panel. We can see that although the three categorical levels are well represented, the points are not nicely spread out in the $(X_1,X_2)$-space. Now consider \texttt{SPlit}. We code the three levels of the categorical variable using Helmert coding. This introduces two dummy variable $d_1$ and $d_2$. Now we find 20 support points in the four-dimensional space and select the subsample using \texttt{SPlit}. They are shown in the right panel of Figure \ref{fig:visual3}. We can see that \texttt{SPlit} picks up three Red and four Blue samples, which agrees with our expectation of choosing approximately $16/100\times 20=3.2$ samples for each of them. The samples are also well-spread out in the $(X_1,X_2)$-space. Thus, clearly \texttt{SPlit} has produced a much better representative sample for testing than the random and stratified proportional subsamples.

\section{Examples}
In this section we will compare \texttt{SPlit} with random subsampling on real datasets for both regression and classification problems.

\subsection{Regression}
Consider the concrete compressive strength dataset from \cite{yeh1998modeling} which can be obtained from the UCI Machine Learning Repository \citep{Dua:2019}. This dataset has eight continuous predictors pertaining to the concrete's ingredients and age. The response is the concrete's compressive strength. We will make an 80-20 split of this dataset which has $1,030$ rows. Thus $N_{train} = 824$ and $N_{test} = 206$. The split is done using both  \texttt{SPlit} and random subsampling. All the nine variables are normalized to mean $0$ and standard deviation $1$ before splitting.

A good splitting procedure should work well for all possible modeling choices. Therefore, to check the robustness against different modeling choices, we choose a linear regression model with linear main effects estimated using LASSO \citep{lasso1996} and a nonlinear-nonparametric regression model estimated using random forest \citep{Breiman2001}. Both the models are fitted on the training set using the default settings of the R packages \texttt{glmnet} \citep{glmnet} and \texttt{randomForest} \citep{randomForest}. Then we compute the root mean squared prediction error (RMSE) on the testing set to evaluate the models' prediction performance. We repeat this procedure 500 times, where the same split is used for fitting both the LASSO and random forest. 

%Obviously, random subsampling generates a new split each time.  Ideally \texttt{SPlit} should give a unique split corresponding to the global minimum of (\ref{eq:support}), but obtaining the global minimum is time consuming and not practical.  The algorithm presented in \citep{mak2018support} for computing support points is designed in such a fashion that every iteration improves the objective function value, and a  reasonable solution is obtained in a few hundred iterations (500 iterations are used here). Owing to random initialization, the solution can vary over different runs of the algorithm. Therefore, the splits obtained using \texttt{SPlit} can also vary in the 500 simulations. 

\begin{figure}[h]
\centering
\includegraphics[width=0.54\columnwidth]{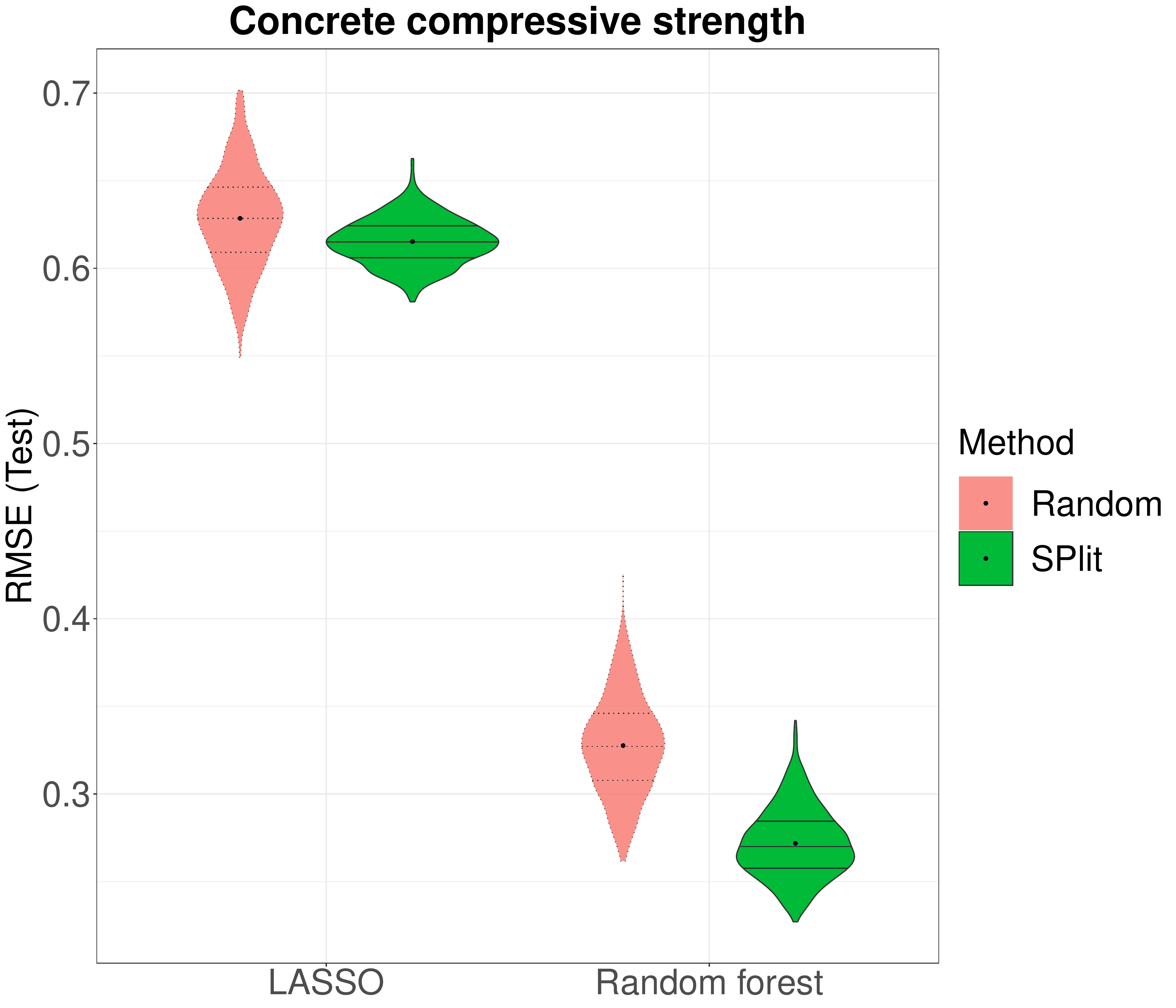}
\caption{Distribution of root mean squared error (RMSE) over 500 random and \texttt{SPlit} subsampling splits for the concrete compressive strength dataset.}
\label{fig:regression}
\end{figure}

The testing RMSE values for the 500 simulations are shown in Figure \ref{fig:regression}. We can see that on the average the testing RMSE is lower for \texttt{SPlit} compared to random subsampling. This improvement is much larger for random forest compared to LASSO. We also note significant improvement in the worst-case performance of \texttt{SPlit} over random subsampling. Furthermore, the variability in the testing RMSE is much smaller for \texttt{SPlit} compared to random subsampling and therefore, a more consistent conclusion can be drawn using \texttt{SPlit}. Thus, the simulation clearly shows that  \texttt{SPlit} produces testing and training set that are much better for model fitting and evaluation. Computation of \texttt{SPlit} for this dataset took on an average $1.6$ seconds on a computer with 6-core $2.6$ GHz Intel processor, which is a negligibly small price that we need to pay for the improved performance over random subsampling.

\begin{table}[h]
\centering
\ra{1.5}
\resizebox{0.75\textwidth}{!}{\begin{minipage}{\textwidth}
\begin{tabular}{lllll}  
\toprule
\textbf{Dataset} & \textbf{Size} & \textbf{Predictors} & \textbf{Response} \\
\midrule
Abalone & $4177 \times 9$ & \makecell[tl]{7 continuous \\ 1 categorical (3 levels)} &  Continuous \\
Airfoil self-noise & $1503 \times 6$ & \makecell[tl]{5 continuous} & Continuous    \\
Meat spectroscopy & $215 \times 101$ & \makecell[tl]{100 continuous} & Continuous   \\
Philadelphia birthweights & $1115 \times 5$ & \makecell[tl]{2 continuous \\ 2 categorical (2 levels each)} & Continuous  \\
\bottomrule
\end{tabular}
\end{minipage}}
\caption{Description of datasets considered for regression.}
\end{table}

\begin{figure}[h]
\centering
\includegraphics[width=0.9\columnwidth]{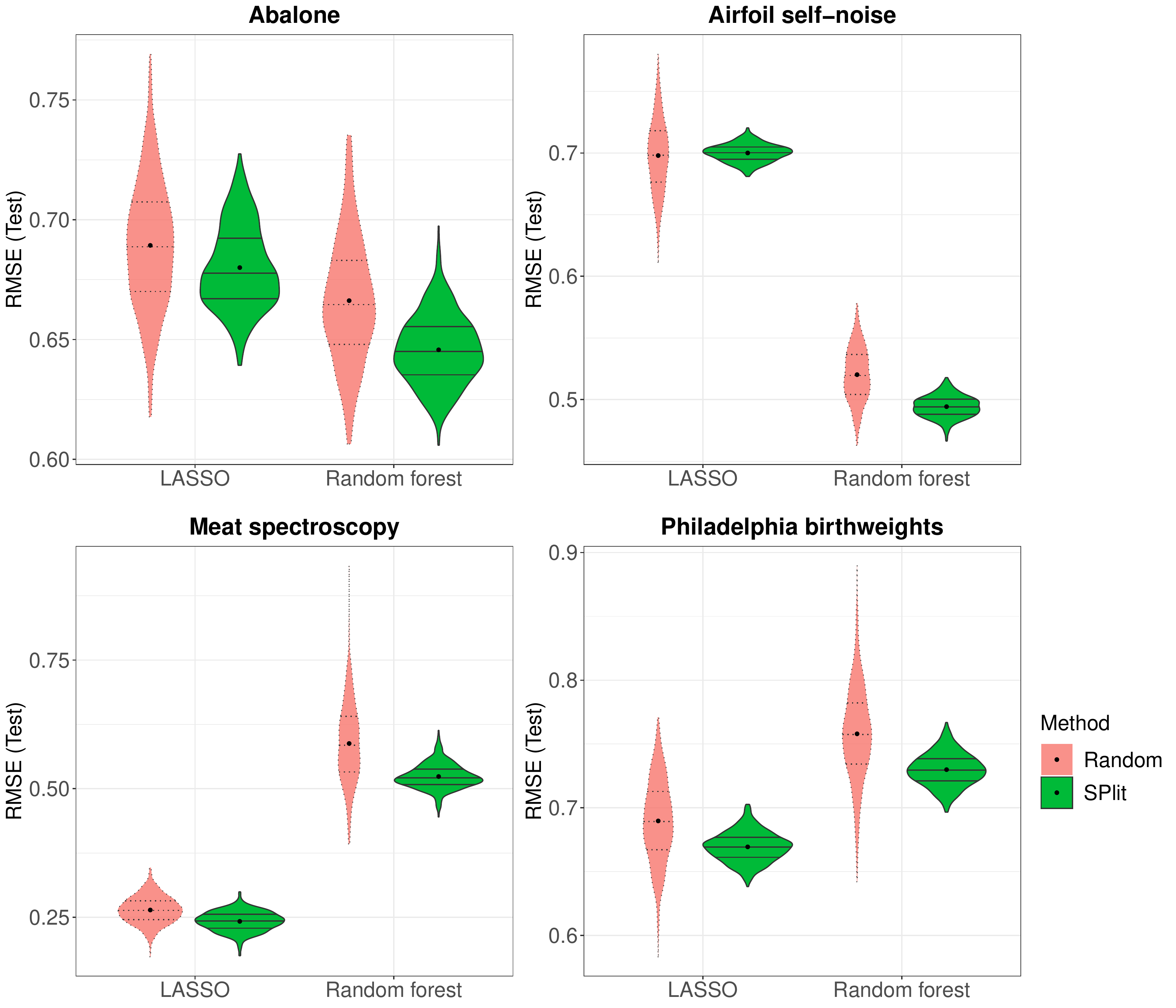}
\caption{Distribution of root mean squared error (RMSE) over 500 random and \texttt{SPlit} subsampling splits for the datasets described in Table 1.}
\label{fig:appendix_regression}
\end{figure}

We repeated the simulation with several other datasets, namely Abalone \citep{nash1994population}, Airfoil self-noise \citep{brooks1989airfoil}, Meat spectroscopy \citep{thodberg1993ace}, and Philadelphia birthweights \citep{elo2001racial}. Abalone and Airfoil self-noise datasets can be obtained from the UCI Machine Learning Repository \citep{Dua:2019}, while Meat spectorscopy and Philadelphia birthweights can be obtained from the \texttt{faraway} \citep{faraway2015} package in \texttt{R}. The details of these datasets are summarized in Table 1. Figure \ref{fig:appendix_regression} shows the testing RMSE values for both LASSO and random forest. We see similar trends as before on all the datasets; \texttt{SPlit} gives a better testing performance on the average than random subsampling and a substantial improvement in the worst-case testing performance.

\subsection{Classification}
For checking the performance of \texttt{SPlit} on classification problems, consider the famous Iris dataset \citep{Fisher1936}. The Iris dataset has four continuous predictors (sepal length, sepal width, petal length, and petal width) and a categorical response with three levels representing the three types of Iris flowers (setosa, versicolor, and virginica). There are 150 rows in total with 50 rows for each flower type. Following the discussion in subsection \ref{sec:coding}, the flower type is converted into two continuous dummy variables using Helmert coding. Thus, the resulting dataset has six continuous columns. For modeling we will use multinomial logistic regression and random forest. The classification performance will be assessed using the residual deviance (\textit{D}) defined as
\begin{equation}
    D := 2 \sum_{i=1}^{I} \sum_{j=1}^{J} y_{ij} \cdot \ln \big(\frac{y_{ij}}{\hat{p}_{ij}}\big) \ ,
\end{equation}
where $I$ is the number of rows, $J$ the number of classes, $y_{ij} \in \set{0, 1}$ is 1 if row $i$ corresponds to class $j$ and 0 otherwise, and $\hat{p}_{ij}$ is the probability that row $i$ belongs to class $j$ as predicted by the model. Note that $0\log 0$ is taken as 0 by definition. 

\begin{figure}[h]
\centering
\includegraphics[width=\columnwidth]{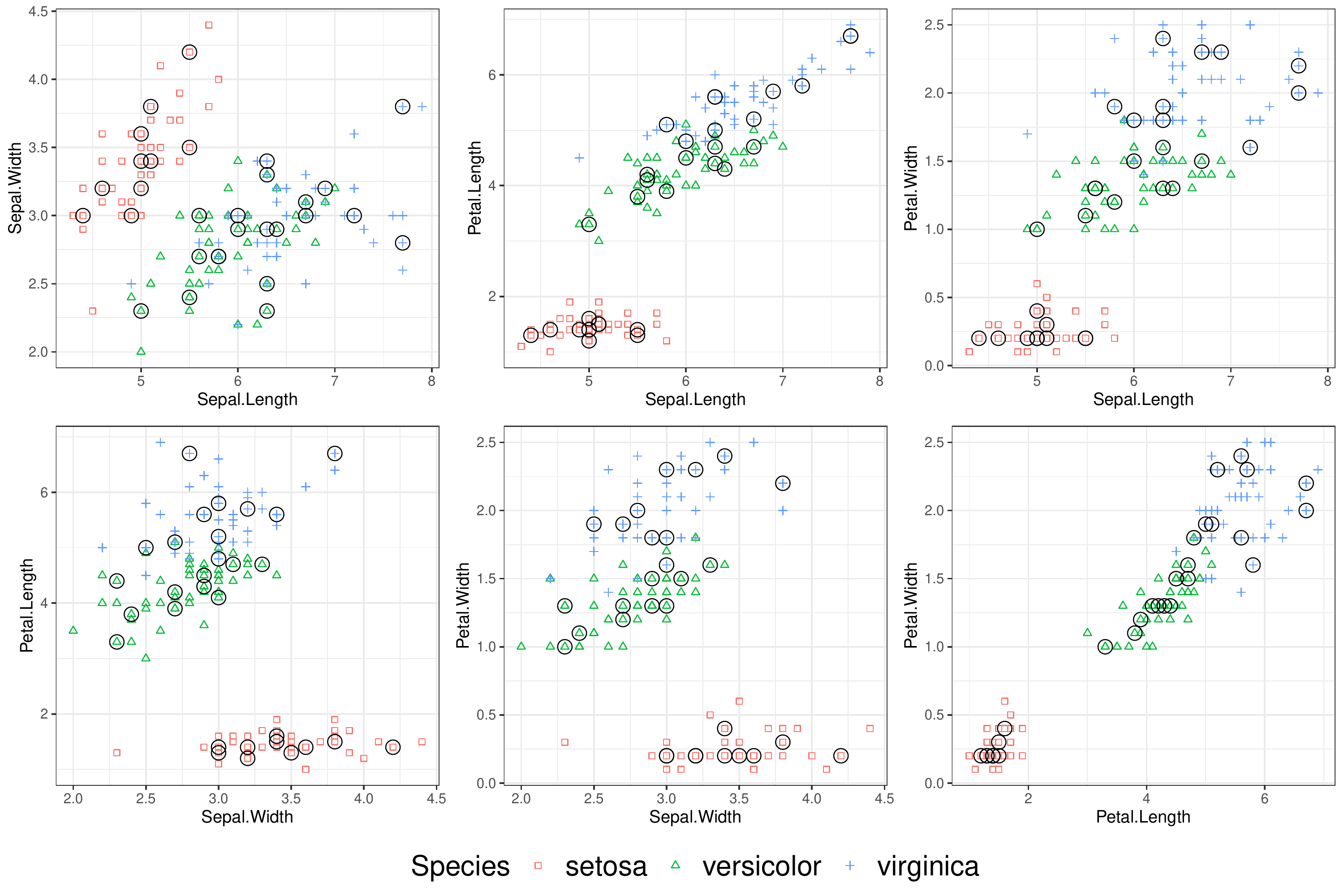}
\caption{Visualizing a \texttt{SPlit} subsampling testing set (circles) for the Iris dataset.}
\label{fig:iris_vis}
\end{figure}

\begin{figure}[H]
\centering
\includegraphics[width=0.6\columnwidth]{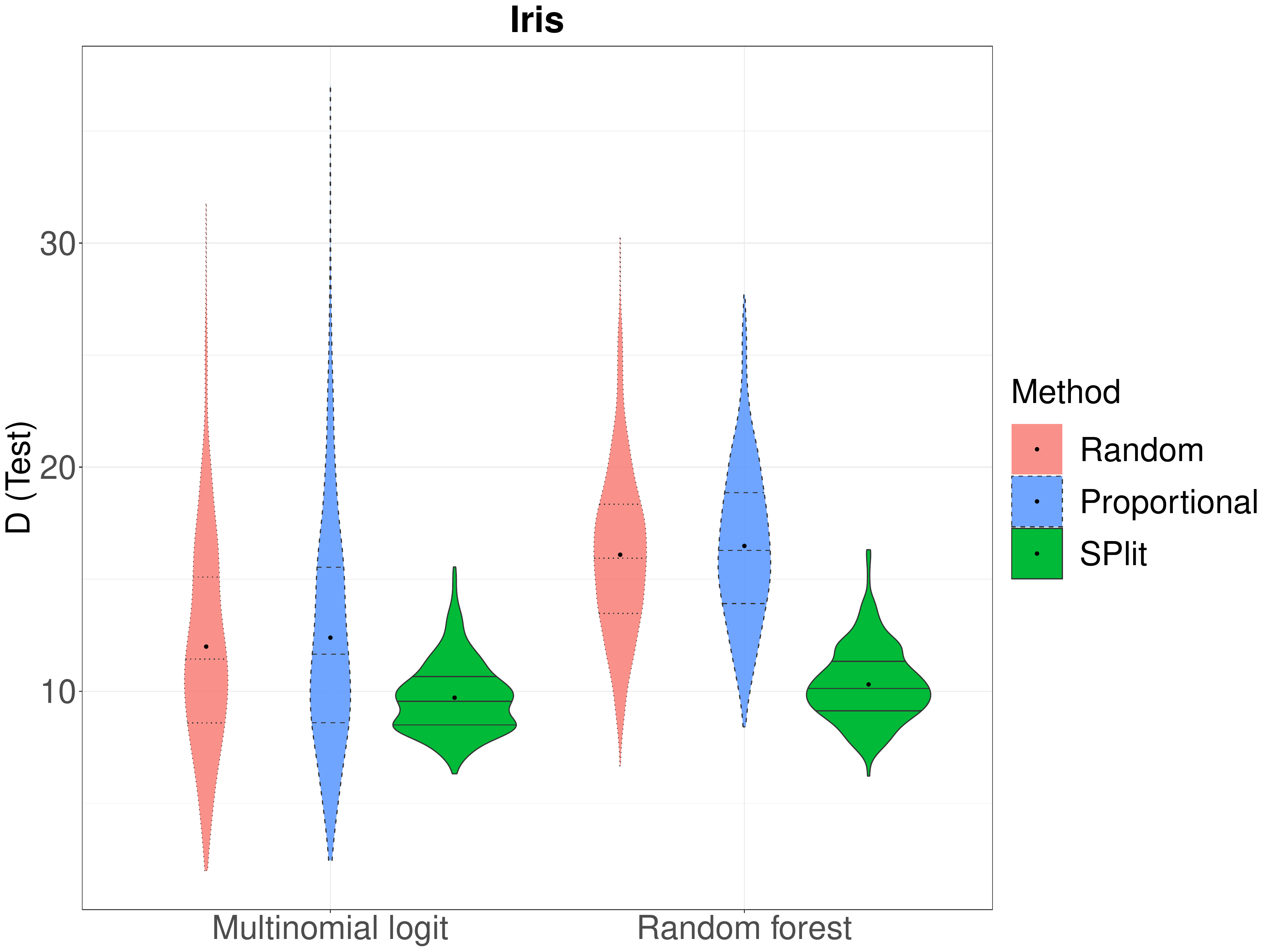}
\caption{Distribution of residual deviance (D) over 500 random, stratified proportional, and \texttt{SPlit} subsampling splits for the Iris dataset.}
\label{fig:classification}
\end{figure}

Figure \ref{fig:iris_vis} shows a testing set selected by \texttt{SPlit}. We can see that they are well-balanced among the three classes and the points are well-spread out in the space of the four continuous predictors. We fit multinomial logistic regression and random forest on the training set and then the residual deviance is computed on the testing set. This is then repeated 500 times. Figure \ref{fig:classification} shows the deviance results for \texttt{SPlit}, random, and stratified proportional subsampling. We can see that again \texttt{SPlit} gives significantly better average and worst-case performance compared to both random and stratified proportional subsampling.

\begin{table}[h]
\centering
\ra{1.5}
\resizebox{0.75\textwidth}{!}{\begin{minipage}{\textwidth}
\begin{tabular}{llll}  
\toprule
\textbf{Dataset} & \textbf{Size} & \textbf{Predictors} & \textbf{Response} \\
\midrule
Banknote authentication & $1372 \times 5$ & \makecell[tl]{4 continuous} &  \makecell[tl]{Categorical (2 levels)}     \\
\makecell[tl]{Breast cancer \\ \textit{(diagnostic, Wisconsin)}} & $569 \times 31$ & \makecell[tl]{30 continuous} & \makecell[tl]{Categorical (2 levels)}    \\
Cardiotocography & $2126 \times 22$ & \makecell[tl]{20 continuous \\ 1 categorical (3 levels)} & \makecell[tl]{Categorical (3 levels)} \\
Glass identification & $214 \times 10$ & \makecell[tl]{9 continuous} & \makecell[tl]{Categorical (6 levels)}  \\
\bottomrule
\end{tabular}
\end{minipage}}
\caption{Description of datasets considered for classification.}
\end{table}

\begin{figure}[H]
\centering
\includegraphics[width=\columnwidth]{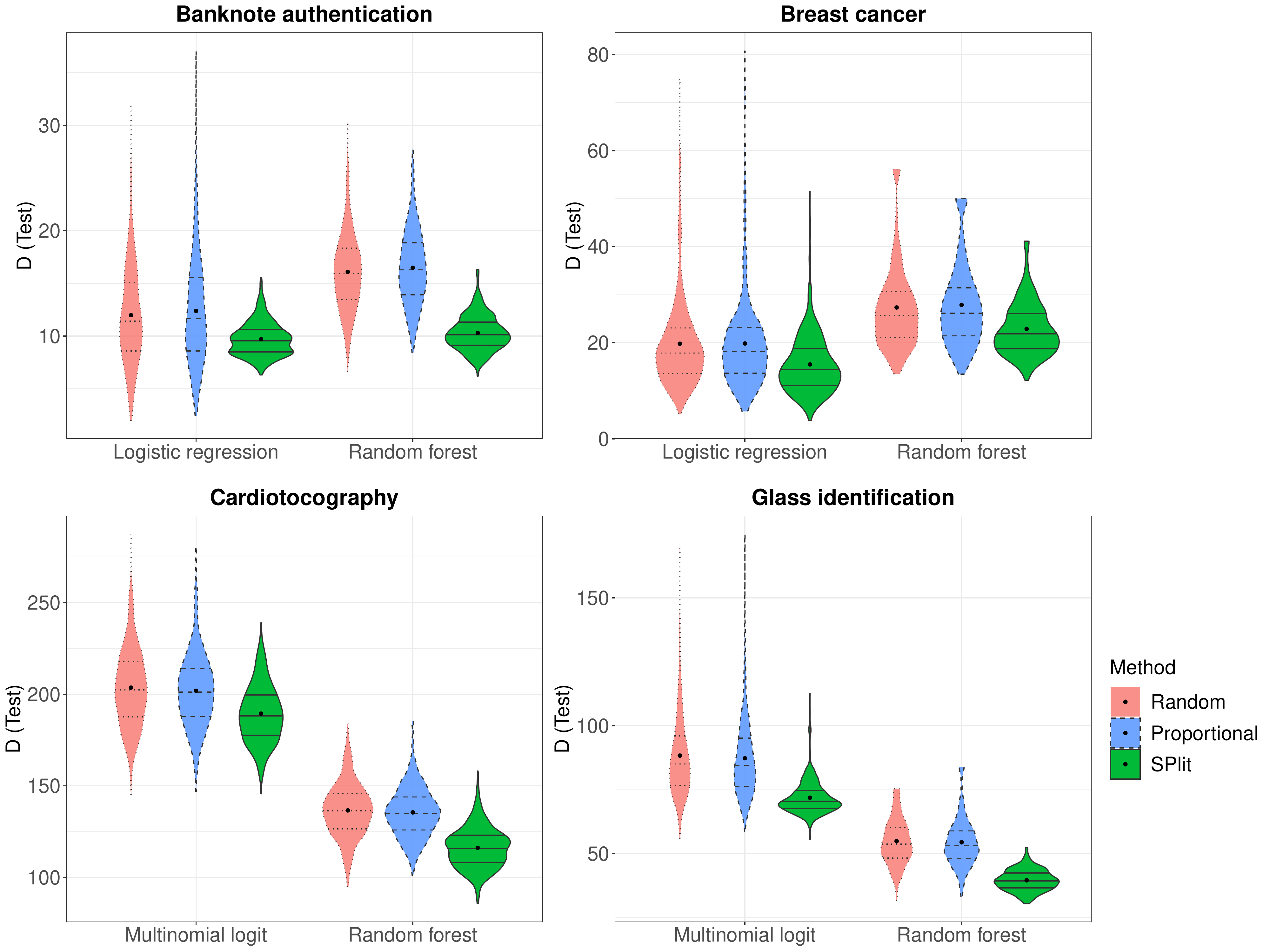}
\caption{Distribution of residual deviance (D) over 500 random, stratified proportional, and \texttt{SPlit} subsampling splits for the datasets described in Table 2.}
\label{fig:appendix_classification}
\end{figure}

The foregoing study is repeated for four other datasets: Banknote authentication, Breast cancer (diagnostic, Wisconsin) \citep{street1993nuclear}, Cardiotocography \citep{ayres2000sisporto}, and Glass identification \citep{evett1989rule}, all of which can be obtained from the UCI Machine Learning Repository \citep{Dua:2019}. The details of these datasets are summarized in Table 2 and the results on the residual deviance are shown in Figure \ref{fig:appendix_classification}. It is possible to encounter $\infty$ while calculating deviance; for the purpose of plotting, $\infty$ is replaced with the maximum finite deviance obtained from the remainder of the 500 simulations. We can see that \texttt{SPlit} gives a better performance than both random and stratified proportional subsampling in all the cases. The improvement realized varies over the datasets and modeling methods, but \texttt{SPlit} has a clear advantage over both random and stratified proportional subsampling.

\section{Conclusions}
Random subsampling is probably the most widely used method for splitting a dataset for testing and training. In this article we have proposed a new method called \texttt{SPlit} for optimally splitting the dataset. It is done by first finding support points of the dataset and then using an efficient nearest neighbor algorithm to choose the subsamples. They are then used as the testing set and the remaining as the training set. The support points give the best possible representation of the dataset (according to the energy distance criterion) and therefore, \texttt{SPlit} is expected to produce a testing set that is best for evaluating the performance of a model fitted on the training set. The ability of support points to match the distribution of the full data is one of its big advantage over the other deterministic data splitting methods such as CADEX and DUPLEX. We have also extended the method of support points  to deal with categorical variables. Thus, \texttt{SPlit} can be applied to both regression and classification problems. We have also briefly discussed on how a sequential application of the support points can be used to generate validation and cross-validation sets, but further development on this topic is left for future research.

We have applied \texttt{SPlit} on several datasets for both regression and classification using different choices of modeling methods and found that \texttt{SPlit} improves the average testing performance in almost all the cases with substantial improvement in the worst-case predictions. The variability in the testing performance metric using \texttt{SPlit} is found to be much smaller than that of random subsampling, which shows that the results and the findings of a statistical study would be much more reproducible if we were to use \texttt{SPlit}.

\begin{center}
    {\Large\bf Acknowledgements}
\end{center}
%We would like to thank Professor Dan Apley for handling this paper as the Editor and for his valuable comments and suggestions. 
This research is supported by a U.S. National Science Foundation grant [CBET-1921873].
%\vspace{.2in}

% \section{Appendix}
% Here we provide the comparison between \texttt{SPlit} and random split for few more datasets. All the models are built in \texttt{R} wherein LASSO and multinomial logistic regression is fit with regularization using the \textit{glmnet} package, and random forest using the \textit{randomForest} package. As before, the splits are 80-20 and the performance metric for regression is root mean squared error (\textit{RMSE}) while for classification it is residual deviance (\textit{D}). 

\bibliography{bibliography}

\end{document}